\documentclass[a4paper,fleqn]{cas-dc}

\usepackage[numbers]{natbib}
\usepackage[ruled,linesnumbered]{algorithm2e}
\usepackage{amsmath}
\usepackage{color}
\usepackage{subfigure}
\usepackage{ulem}
\usepackage{arydshln} 
\usepackage{booktabs}
\usepackage{multirow}
\usepackage{makecell} 
\usepackage{booktabs}
\usepackage{multicol}
\usepackage{tabu} 
\usepackage{hhline}
\usepackage{multirow}
\usepackage{diagbox}
\usepackage{cancel}
\usepackage{caption}
\usepackage{booktabs, tabularx}
\usepackage{stfloats}
\usepackage{caption}
\usepackage{supertabular}
\usepackage{lipsum} 

\newcommand\cb{\color{black}} 
\newcommand\cbt{\color{black}} 

\newcommand{\tabincell}[2]{\begin{tabular}{@{}#1@{}}#2\end{tabular}}
\usepackage{amsfonts, amsmath, mathrsfs,  amssymb, dsfont, mathtools, pifont, amsbsy, graphicx,  yfonts, bbm, bm}

\def\({\left(}
\def\){\right)}
\def\[{\left[}
\def\]{\right]}
\def\BEq{\begin{eqnarray}}
\def\EEq{\end{eqnarray}}
\def\BE*{\begin{eqnarray*}}
\def\EE*{\end{eqnarray*}}
\def\BA{\begin{array}}
\def\EA{\end{array}}

\def\bTheta{\mathbf{\Theta}}

\def\bLambda{\mathbf{\Lambda}}

\def\0{\mathbf{0}}
\def\1{\mathbf{1}}

\def\A{\mathbf{A}}

\def\c{\mathbf{c}}

\def\D{\mathbf{D}}

\def\g{\mathbf{g}}

\def\I{\mathbf{I}}

\def\l{\mathbf{l}}

\def\q{\mathbf{q}}

\def\bR{\mathbb{R}}

\def\U{\mathbf{U}}

\def\W{\mathbf{W}}

\def\X{\mathbf{X}}

\def\Y{\mathbf{Y}}

\def\and{\prefixtext{and}}

\begin{document}
\let\WriteBookmarks\relax
\def\floatpagepagefraction{1}
\def\textpagefraction{.001}

\shorttitle{{\cb POCKET}: Pruning Random Convolution Kernels for Time Series Classification from a Feature Selection Perspective}

\shortauthors{Shaowu Chen et~al.}

\title [mode = title]{{\cb POCKET}: Pruning Random Convolution Kernels for Time Series Classification from a Feature Selection Perspective}



%

\author[1,2]{Shaowu Chen}[orcid=0000-0001-7511-2910]
\ead{shaowu-chen@foxmail.com}
\credit{Conceptualization, Methodology, Software, Visualization, Writing - Original draft preparation}

\author[1]{Weize Sun}[orcid=0000-0002-8658-7775]
\ead{proton198601@hotmail.com}
\cormark[1]
\credit{Methodology, Funding acquisition, Formal analysis, Resources, Project administration, Writing - review \& editing}

\author[1]{Lei Huang}[orcid=0000-0003-4534-7625]
\ead{lhuang@szu.edu.cn}
\credit{Methodology, Funding acquisition, Resources, Supervision, Writing - review \& editing}

\author[1]{Xiao Peng Li}[orcid=0000-0003-2503-690X]
\ead{x.p.li@szu.edu.cn}
\credit{Conceptualization, Resources, Writing - review \& editing}

\author[2]{Qingyuan Wang}[orcid=0000-0002-7879-4328]
\ead{qingyuan.wang@ucdconnect.ie}
\credit{Validation, Software, Writing - review \& editing}

\author[2]{Deepu John}[orcid=0000-0002-6139-1100]
\ead{deepu.john@ucd.ie}
\credit{Conceptualization, Resources, Writing - review \& editing}

\affiliation[1]{organization={State Key Laboratory of Radio Frequency Heterogeneous Integration (Shenzhen University)},
    addressline={Shenzhen, 518060},
    country={China}
    }

\affiliation[2]{organization={School of Electrical and Electronic Engineering, University College Dublin},
    city={Dublin 4},
    postcode={D04 V1W8}, 
    country={Ireland}
    }

\cortext[cor1]{Corresponding author}



\begin{abstract}
In recent years, two competitive time series classification models, namely, ROCKET and MINIROCKET, have garnered considerable attention due to their low training cost and high accuracy. However, they {\cbt rely on} a large number of random 1-D convolutional kernels to comprehensively capture features, which is incompatible with resource-constrained devices. Despite the development of heuristic algorithms designed to recognize and prune redundant kernels, the inherent time-consuming nature of evolutionary algorithms hinders efficient evaluation.
{\cbt
To efficiently prune models, this paper eliminates feature groups contributing minimally to the classifier, thereby discarding the associated random kernels without direct evaluation. To this end, we incorporate both group-level ($l_{2,1}$-norm) and element-level ($l_2$-norm) regularizations to the classifier, formulating the pruning challenge as a group elastic net classification problem. An ADMM-based algorithm is initially introduced to solve the problem, but it is computationally intensive. Building on the ADMM-based algorithm, we then propose our core algorithm, POCKET, which significantly speeds up the process by dividing the task into two sequential stages. In Stage 1, POCKET utilizes dynamically varying penalties to efficiently achieve group sparsity within the classifier, removing features associated with zero weights and their corresponding kernels. In Stage 2, the remaining kernels and features are used to refit a $l_2$-regularized classifier for enhanced performance.}
Experimental results on diverse time series datasets show that POCKET prunes up to 60\% of kernels without a significant reduction in accuracy and performs 11$\times$ faster than its counterparts. 
Our code is publicly available at \href{https://github.com/ShaowuChen/POCKET}{ https://github.com/ShaowuChen/POCKET}.

\end{abstract}




\maketitle

\section{Introduction}

The time series classification (TSC) task has been widely explored in diverse application domains, including electrocardiogram diagnosis \cite{ECG}, incident detection \cite{ROCKETCaseStudy}, and fault locating \cite{DLFaultDiagnosis}. In recent years, numerous advanced algorithms or models for TSC have emerged \cite{ismail2020inceptiontime, 10129257, sun2023ranking, salehinejad2017recent} to improve accuracy.  
However, a discernible trend is that the number of training parameters of algorithms is constantly increasing, incurring significant training time and computational resource overhead. This phenomenon becomes particularly conspicuous when dealing with large datasets.

The high complexity of various TSC methods has sparked an exploration of faster and more scalable alternatives \cite{REVIN2023110483, lucas2019proximity, cboss}. The representative state-of-the-art works are ROCKET (for \textbf{R}and\textbf{O}m \textbf{C}onvolutional \textbf{KE}rnel \textbf{T}ransform) \cite{dempster2020rocket} and MINIROCKET \cite{dempster2021minirocket}, which leverage random 1-D convolutional kernels to extract features, namely, the proportion of positive values (PPV) and/or maximum values (MAX), which are then used to train linear classifiers.
As the random kernels are training-free, ROCKET and MINIROCKET are more efficient in fitting classifiers compared to related TSC methods. Moreover, the utilization of PPV and MAX also reduces the dimension of the input data, reducing the complexity of the classifiers.

However, due to the randomness of convolution kernels, the extracted features may not be sufficiently discriminative for accurate classification. To comprehensively capture the temporal characteristics of time series data, ROCKET and MINIROCKET employ a large number of kernels, typically 10,000. Nevertheless, such an abundance of kernels introduces significant time and memory overhead during the inference process, making it challenging to deploy ROCKET and MINIROCKET on resource-constrained devices for real-time tasks. Additionally, the extracted features show a notable degree of redundancy \cite{salehinejad2022s}, indicating that a substantial proportion of randomly generated kernels have a negligible impact on the TSC outcomes. Therefore, it becomes imperative to identify and prune redundant random kernels to enhance the efficiency of ROCKET and MINIROCKET while simultaneously preserving classification performance.

{\cb
Selecting the optimal subset of kernels from the kernels of ROCKET and MINIROCKET can be considered a subset selection or a combinatorial optimization problem, whose goal is to maximize the accuracy of models while minimizing the number of kernels.
Directly selecting kernels is NP-hard, due to the exponential number of possible combinations of kernels as the number of kernels increases, which makes it computationally infeasible to evaluate every possible subset to determine the optimal one. }
{\cb Additionally,  as the random kernels are training-free,} 
data-independent criteria such as $\ell_1$ norm \cite{0022KDSG17} and WHC \cite{chen2023whc} are less applicable.
To tackle this challenge, \citet{salehinejad2022s} introduced a differential evolutionary algorithm named S-ROCKET to effectively delete redundant kernels, albeit with a slight performance degradation. However, the evaluation period of S-ROCKET is time-consuming. Another potential drawback is that the pruning rate is determined by the evolutionary algorithm and cannot be predetermined based on available computational resources. This increases the risk of redundancy removal failure, where 100\% of the kernels are retained, as observed in our experiment \ref{Comparing with current state of the art}.

\begin{figure*}[t]
\centering
\includegraphics[width=0.8\linewidth]{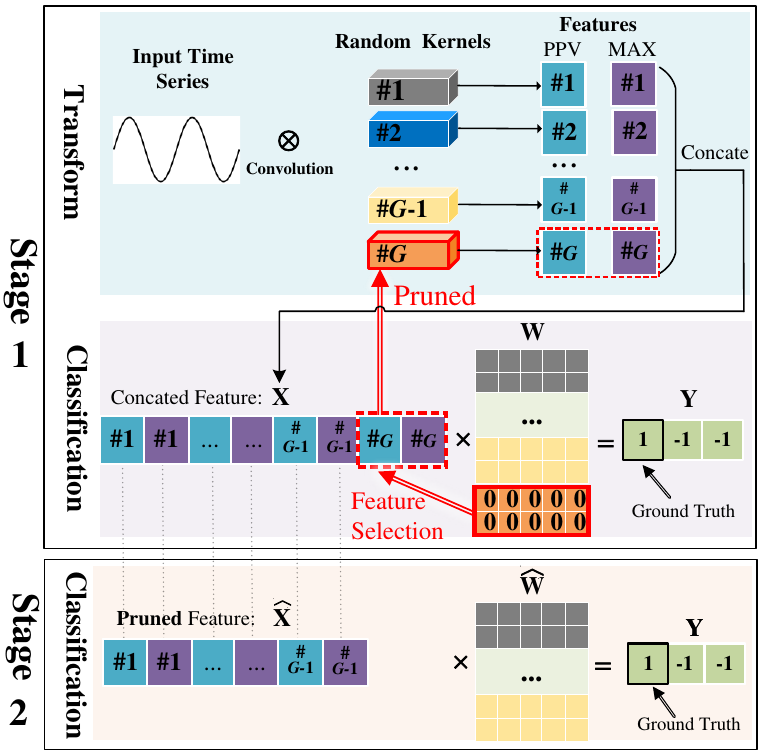}
\caption{{\cb Overall diagram of POCKET. \textbf{TOP:} Stage 1, {\cbt remove feature groups that minimally contribute to the $\ell_{2,1}$-regularized classifier, thereby eliminating the corresponding kernels}. \textbf{BOTTOM:} (optional) Stage 2, refit a $\ell_2$-regularized classifier using the remaining (kernels and) features for enhanced performance. }}
\label{model}
\end{figure*}

In contrast to existing approaches such as \cite{salehinejad2022s, AccROCKETWave}, which directly select or modify kernels, this paper {\cb proposes to} efficiently prune the `ROCKET family' through feature selection at the classification layer.  The motivation behind this is that it is difficult to identify redundant kernels, as random kernels generated from the identical distribution do not exhibit significant differences, while the extracted features play notably distinguishable roles in the classification layer.
As illustrated {\cb at the top of} Figure \ref{model}, if all the features extracted by a kernel minimally contribute to the subsequent classifier, the kernel can be considered redundant and safely removed without a substantial drop in accuracy.
To achieve this, we formulate the pruning task as a $\ell_{2,1}$ constrained classification problem and incorporate $\ell_2$ regularization to enhance model generalization, thereby transforming the challenge into a group elastic net problem. To solve this problem, we propose an ADMM-based algorithm that utilizes a dynamically varying soft threshold \cite{liao2014generalized} during iterations, enabling a user-defined pruning rate. However, the iterative calculation of the inverse of a large matrix in each iteration makes the ADMM-based algorithm time-consuming.

To accelerate the solving process, we further introduce our core algorithm, \textbf{{\cb POCKET}}, {\cb which stands for} "\textbf{P}ruing R\textbf{OCKET}s". Building on the initial ADMM-based algorithm, {\cb POCKET} divides the $l_{2,1}$ and $l_2$ regularizations into two consecutive stages, solving them separately. 
{\cb As depicted at the top of Figure \ref{model},} Stage 1 of {\cb POCKET} prunes random kernels via feature selection by fitting a group-sparse classifier. It introduces relatively invariant penalties to avoid calculating the inverse in each iteration, significantly reducing the time complexity for achieving a solution. Stage 2{\cb, shown at the bottom of Figure \ref{model},} is an optional phase that implements element-wise regularization to refit a high-accuracy linear classifier using the remaining features from Stage 1. In our experiments, Stage 2 {\cb effectively mitigates} the performance degradation of {\cb pruned} MINIROCKET, which is less redundant and therefore sensitive to pruning.

The main contributions of this paper are summarized as follows.
\begin{enumerate}
\item{We introduce a novel approach to pruning ROCKET models from a feature selection perspective, providing a more feasible alternative compared to the existing methods that directly evaluate random kernels.}
\item{The challenge of kernel pruning is addressed by formulating it as a group elastic net problem. We propose two algorithms to solve this problem, with the introduced {\cb POCKET} showcasing an efficiency that is 11$\times$ faster than its counterparts. {\cb POCKET} also allows for adjustable pruning rates in accordance with the computational budget.}
\item{Extensive experiments across various datasets are conducted to thoroughly evaluate the performance and efficacy of the proposed algorithms.
{\cb The results demonstrate that when more than 60\%  of kernels on average are removed, the accuracy of ROCKET pruned by POCKET does not decrease significantly, but instead improves slightly.}}

\end{enumerate}

The remainder of this paper is organized as follows. In Section \ref{RelatedWork}, related works are summarized. Section \ref{Methodology} formulates the problem and introduces the proposed algorithms. In Section \ref{Experiments}, experimental results are reported and discussed. Section \ref{Conclusion} concludes the paper.


\section{Related Work}
\label{RelatedWork}

\subsection{Time Series Classification}
As TSC deals with inherently ordered data, algorithms for TSC are predominantly designed to capture distinctive order patterns of data, enabling effective classification. We categorize these approaches into six interrelated categories as outlined below, excluding the `ROCKET family', which is elaborated on in the subsequent Subsection \ref{Random Kernel Based ROCKET and MINIROCKET}. For a more comprehensive introduction to TSC,  readers can refer to relevant reviews~\cite{BagnallLBLK17, ruiz2021great}.

\textbf{Similarity based.} This category usually leverages alignment metrics to measure similarity between time series data.
{\cbt
To tackle the time-mismatch problem caused by various lengths of series in p-norm measure,} elastic measures have emerged as promising alternatives to mitigate these limitations. For example, dynamic time warping (DTW) \cite{shokoohi2017generalizing, jeong2011weighted}, complexity-invariant distance \cite{batista2014cid}, and elastic ensemble \cite{lines2015time} can evaluate the similarity of unaligned series and provide higher accuracy. 
{\cbt Rather than calculating distance, pattern-based algorithms evaluate the similarity of trends or relative orders \cite{OPPMiner, OPRMiner}. Motivated by OPR-Miner \cite{OPRMiner}, which mines frequent patterns for classification, \citet{COPPMiner} proposed COPP-Miner to discover the most contrasting patterns, which are then used to train classifiers with superior performance.
}

\textbf{Interval based.} Rather than utilizing the entire time series for classification, interval-based methods split the time series into numerous intervals.
The Time Series Forest (TSF) \cite{deng2013time} constructs decision trees based on the statistical attributes of randomized intervals of the time series.  In contrast, Random Interval Forest \cite{lines2016hive} selects random intervals and extracts spectral features, avoiding the use of statistical features. 
{\cbt
Using aggregated features extracted from randomly selected intervals,
the randomized-supervised time series forest (r-STSF) \cite{cabello2024fast} performs orders of magnitude faster than ensemble-based algorithm while being competitive in accuracy. }

\textbf{Shapelet based.} Shapelet is a representative subsequence that encapsulates the fundamental characteristics of a dataset. Shapelet-based approaches assess the similarity between time series data and shapelets for TSC. \citet{ye2011time} enumerated all conceivable candidates as shapelets and used an embedded decision tree structure. To accelerate the discovery of shapelets, the Fast Shapelets approach \cite{rakthanmanon2013fast} uses a randomized projection technique to represent the symbolic aggregate approximation (SAX) of data. The Shapelet Transform \cite{hills2014classification} leverages optimal shapelets to evaluate distances and facilitate the training of classifiers. Contrastingly, \citet{grabocka2014learning} applied a gradient descent algorithm to a regularized logistic loss to learn shapelets, which are not present in the original series but are effective for TSC.

\textbf{Dictionary based.} Shapelet-based techniques may fail when class differentiation hinges on the relative occurrence of patterns rather than local subsequences. In such a scenario, dictionary-based methodologies \cite{dempster2023hydra, tabassum2022time} demonstrate efficacy. The Bag-of-SFA-Symbols (BOSS) algorithm \cite{schafer2015boss} converts subsequences into `words' using sliding windows and subsequently tally the frequency of recurring words. An evolution of BOSS, known as cBOSS \cite{cboss}, introduces a scalable variant. By amalgamating various dictionary methods, including cBOSS, the Temporal Dictionary Ensemble (TDE) \cite{middlehurst2021temporal} achieves state-of-the-art accuracy. 

\textbf{Ensemble based.}
Ensemble-based approaches use various classifiers to enhance performance. The HIVE-COTE algorithm \cite{lines2018time} classifies data using a weighted average of multiple algorithms, including BOSS, Shapelet Transform, elastic distance measure, and frequency representation classifiers. Recently, various HIVE-COTE variants {\cbt adapting TDE \cite{middlehurst2021temporal}, Canonical Interval Forest (CIF) \cite{middlehurst2020canonical}, and Diverse Representation CIF \cite{middlehurst2021hive} } have developed and shown significant improvement. 

\textbf{Deep Learning Based.}
The advancements in deep learning also contribute to the ongoing development of TSC \cite{zuo2023svp, XIAO2021106934,foumani2024improving, ismail2019deep, wen2023transformers}. 
InceptionTime \cite{ismail2020inceptiontime} uses cascading inception modules \cite{szegedy2016rethinking} to extract important features from time series subsequences of various lengths.
{\cbt
By combining hand-crafted filters with InceptionTime,
H-ITime \cite{H-IT2022} captures discriminate patterns across classes and achieves enhanced performance.
To reduce the number of parameters, L-Time \cite{LITE2023} employs custom convolutions and proposes a light-weight variant of InceptionTime.}

\subsection{Random Kernel Based Models}
\label{Random Kernel Based ROCKET and MINIROCKET}
Another type of TSC method is the `ROCKET family' based on random kernels, including ROCKET\cite{dempster2020rocket}, MINIROCKET \cite{dempster2021minirocket}, and their variants such as MultiROCKET \cite{tan2022multirocket}.

The pioneering ROCKET model developed by \citet{dempster2020rocket} utilizes a large number of 1-D random convolutional kernels, defaulting to 10,000, without the need for training. These kernels transform input time series into two sets of features, namely {\cb PPV} and {\cb MAX}, resulting in a total of 20,000 features, as depicted in Figure \ref{model}. To fully capture features, the random kernels employed in ROCKET are parameterized with varying lengths, biases, dilations, and padding. Subsequently, a linear classifier is trained on the extracted features for classification, achieving state-of-the-art performance \cite{tan2021time}. As the random kernels are train-free and only the classifier is fitted, the ROCKET model can efficiently train in a very short time.

MINIROCKET \cite{dempster2021minirocket} is an improved variant of ROCKET, marked by two key distinctions. Firstly, MINIROCKET employs (almost) deterministic random kernels with binary values, such as 0 and 1, significantly accelerating convolution on samples. Secondly, MINIROCKET only utilizes the PPV feature, as the MAX feature is deemed less influential on classification accuracy. Consequently, the number of features is halved from 20,000 to 10,000. These enhancements also render MINIROCKET up to 75 times faster than ROCKET while maintaining competitive accuracy compared to other methods

Recently, a series of works building upon or drawing inspiration from ROCKET and MINIROCKET \cite{tan2022multirocket, dempster2023hydra, schlegel2022hdc} have emerged. A common thread among these works is their continued reliance on a substantial number of random kernels to preserve the diversity of transformed features, albeit with potentially varying kernel configurations. Consequently, similar to ROCKET and MINIROCKET, these methodologies also contend with the challenge of excessive redundancy among random kernels, presenting obstacles to practical deployment. 

Despite variations in the configurations of random kernels across different models, our proposed {\cb POCKET} remains an effective approach for reducing redundancy. By disregarding the specifics of the kernels themselves and concentrating on pruning through feature selection within the classification layer, {\cb POCKET} provides a generic solution to the common issue of kernel redundancy observed in various models relying on random kernels.

\subsection{Model Compression and S-ROCKET}
Model compression has attracted significant attention in recent years, particularly for convolutional neural networks (CNNs). Four prominent strategies for compressing CNNs include low-rank factorization \cite{LIU2022109723, sun2020deep, chen2023joint,LIU2022108171}, quantization \cite{lyu2023designing}, neural architecture search \cite{huang2023split}, and pruning \cite{LEE2022107988, CHANG2023110386,preet2022class}. By considering the computation of PPV and MAX from convolutional feature maps as special pooling operations, ROCKET and MINIROCKET can be regarded as specialized shallow CNNs, each consisting of only one random hidden layer and one pooling layer. Therefore, it is reasonable to explore compressing ROCKET and MINIROCKET using CNN compression techniques, but the randomness of kernels makes the borrowing less straightforward. 

Several works have specifically focused on pruning redundant random kernels within the `Rocket family' \cite{salehinejad2022s, AccROCKETWave, 10095893}. S-ROCKET \cite{salehinejad2022s} is the representative pioneer that can prune around 60\% of kernels with minor performance degradation following the three-step pipeline: pre-training, kernel selection, and post-training. However, the evolutionary algorithm used in the second step for kernel selection demands extensive iterations for evaluations and introduces a notable time overhead. Additionally, the pruning rate of S-ROCKET depends on the evaluation, introducing the potential risk of failure in pruning if a 0\% pruning rate is encountered, as observed in our experiments in Section \ref{exp-b-2}. In contrast,  our proposed algorithms can achieve customized pruning rates according to the computational budget.

\section{Methodology}
\label{Methodology}
In \ref{notationis} and \ref{Problem Formulation}, we define notations and formulate the problem, respectively. Subsequently, our first ADMM-based algorithm is introduced in \ref{Solution using ADMM} to solve the problem. In \ref{accPOCKET}, taking the ADMM-based algorithm as a prelude, we introduce our core two-stage algorithm, {\cb POCKET}, to significantly accelerate the {\cb pruning} process.

\subsection{Notations}
\label{notationis}
In this paper, italicized,  bold lowercase, and bold uppercase letters are used to represent scalars, vectors, and matrices, respectively. 
The notation $\I_d$ represents the identity matrix of shape $d\times d$.
The $i$-th row, ($i,j$)-th entry and trace of a matrix $\A$ are denoted as $\A_{i:}$, $\A_{i,j}$ and $\text{tr}(\A)$, respectively.

For simplicity, here we define the reshaped form of a matrix $\mathbf{A} \in \bR^{I\times J}$ as $\mathbf{A}'$ with an operator $\text{reshape}(\cdot)$:
\begin{align}
\label{reshape}
\A':&=\text{reshape}(\A)
\\
&= 
\begin{cases}
\ \A \in \bR^{I\times J},  \qquad \qquad\text{in MINIROCKET,} \\
\\
\
\begin{bmatrix}
\A_{1:},&\A_{2:}\\
\A_{3:},&\A_{4:}\\
\cdots\\
\A_{I-1:},&\A_{I:}\\
\end{bmatrix} \in \bR^{\frac{I}{2}\times2J},     \ \text{in ROCKET.}\\
\end{cases}\nonumber
\end{align}

\subsection{Problem Formulation}
\label{Problem Formulation}

Different from S-ROCKET \cite{salehinejad2022s}, which focuses on the convolutional layer with random weights,  we aim to prune redundant kernels effectively by eliminating related features in the classifier, as illustrated in Figure \ref{model}.
To this end, we impose group-level or row-wise sparsity in the classifiers for ROCKET and MINIROCKET, respectively:
\begin{align}
\label{p0}
\begin{split}
    &\min_{\W} \quad \Vert \W' \Vert_{2,0} \\
     \text{s.t.}& \quad  \Vert \W\Vert_F < c_1, \quad \X\W=\Y.
\end{split}
\end{align}
Here, $\W \in \mathbb{R}^{{\cb H} \times C}$ is the weight matrix for $C$-class classification, $\W'=\text{reshape}(\W)$ as defined in equation (\ref{reshape}), and $c_1$ is a user-defined constant that limits the magnitude of $\W$ and enhances the generalization of the classifier.
$\X\in \mathbb{R}^{N\times {\cb H}}$ represents the feature matrix constituted by ${\cb H}$ features extracted by
${\cb G}$ random convolution kernels from $N$ samples, where ${\cb G}:={\cb H}$ for MINIROCKET and ${\cb G}:=\frac{{\cb H}}{2}$ for ROCKET, respectively. (Note that each kernel of ROCKET extracts PPV and MAX, leading to two consecutive rows within $\mathbf{X}$ originating from a single kernel; whereas MINIROCKET solely  extracts PPV.) Here $\Y \in \mathbb{R}^{N\times C}$ corresponds to the class indicator. The ground-truth $\Y$ is binary-coded ($\pm 1$), so that the only positive element in each row indicates the class of the corresponding time series data.
This manner effectively achieves $C$-class classification using $C$ two-category classifiers based on regression. 
During the prediction phase, $\X\W$ is usually not strictly equal to the ground-truth $\Y$, and the categories of samples are indicated by the indices of the largest elements within each row of $\X\W$.

Solving the $\ell_{2,0}$ problem (\ref{p0}) is NP-hard, therefore, we reformulate our objective as a group elastic net problem:
\begin{align}
\label{p1}
\begin{split}
   &\min_{\W}  \quad \Vert \W' \Vert_{2,1}\\
    \text{s.t.} \quad &\Vert \W\Vert_F<c_1, \quad \Vert\X\W-\Y\Vert_F< c_2,
\end{split}
\end{align}
which is convex and easier to solve. Here, $\Vert \W' \Vert_{2,1}$ is the convex approximation of $\Vert \W' \Vert_{2,0}$. We substitute the strict constraint $\X\W=\Y$ with the relaxed $\Vert\X\W-\Y\Vert_F< c_2$, where $c_2$ is a hyperparameter, to improve the noise tolerance of the classifier.

\subsection{Prelude: ADMM-based Algorithm}
\label{Solution using ADMM}
In this subsection, we propose an ADMM-based algorithm to solve the problem (\ref{p1}). Based on the algorithm, we will introduce the core two-stage pruning algorithm, the accelerated {\cb POCKET}, in the next subsection.

To utilize ADMM \cite{LEE2022107988, boyd2011distributed}, we rewrite problem (\ref{p1}) as
\begin{align}
\label{p2}
    \min_{\W,\bTheta}& \quad \Vert \bTheta' \Vert_{2,1} \nonumber\\
    \text{s.t.}& \quad \Vert \W\Vert_F<c_1,\\
     &\quad \Vert\X\W-\Y\Vert_F< c_2,\nonumber\\
     &  \quad \bTheta=\W,\nonumber
\end{align}
where $\bTheta'=\text{reshape}({\bTheta})$. Here and in the subsequent sections, we assume that rows of $\X$ are zero-centered and scaled to have unit $\ell_2$ norms, and the $\Y$ is column-wise zero-centered.
By applying the penalty method to the inequality constraints and attaching the Lagrangian multiplier to the equality constraint, we formulate the augmented Lagrangian function for the problem (\ref{p2}) as
\begin{align}
\text{L}\left(\W,\bTheta, \bLambda\right)&=\Vert\bTheta'\Vert_{2,1}+\frac{\rho_1}{2}\Vert\W\Vert_F^2+\frac{\rho_2}{2}\Vert \X\W-\Y\Vert_F^2\nonumber\\
&+\frac{\rho_3}{2}\Vert\bTheta-\W\Vert_F^2+\text{tr}\left(\bLambda^T \circ (\bTheta-\W)\right),
\end{align}
where $\bLambda\in\mathbb{R}^{{\cb H}\times C}$ denotes the Lagrange multiplier, and `$\circ$' signifies element-wise multiplication. {\cb Here $\rho_1$, $\rho_2$, and $\rho_3$ are positive penalty scalars, which are user-defined hyperparameters. Generally, a smaller value of $c_1$ leads to a larger $\rho_1$, and the same relationship holds for $c_2$ and $\rho_2$.}

Leveraging ADMM \cite{boyd2011distributed},  problem (\ref{p2}) can be solved by iteratively updating the following variables:
\begin{align}
\label{updateW}
    &\W^{(t)}= \arg \min_{\W}\quad \text{L}\left(\W,\bTheta^{(t-1)}, \bLambda^{(t-1)}\right),\\
\label{updateTheta}
    &\bTheta^{(t)}= \arg \min_{\bTheta} \quad\text{L}\left(\W^{(t)},\bTheta, \bLambda^{(t-1)}\right),\\
    &\bLambda^{(t)}= \bLambda^{(t-1)}+ s * \left(\bTheta^{(t)}-\W^{(t)}\right)\label{dual},
\end{align}
where $t$ signifies the step of iteration, and $s>0$ is the step size for updating the multiplier. 
 
\subsubsection{Updating $\W$}

By defining 
\begin{align}
    \U = \frac{\bLambda}{\rho_3},\label{defineU}
\end{align}
the problem (\ref{updateW})  can be  formulated as
\begin{small}
\begin{align}
\W^{(t)}&=\arg \min_{\W}\frac{\rho_1}{2}\Vert\W\Vert_F^2+\frac{\rho_2}{2}\Vert \X\W-\Y\Vert_F^2\nonumber\\
&+\frac{\rho_3}{2}\Vert\bTheta^{(t-1)}-\W\Vert_F^2+\text{tr}\left(\{\bLambda^{(t-1)}\}^T \circ (\bTheta^{(t-1)}-\W)\right)\nonumber\\
    &=\arg \min_{\W} \frac{\rho_1}{2}\Vert\W\Vert_F^2+\frac{\rho_2}{2}\Vert \X\W-\Y\Vert_F^2 \nonumber\\
    &\quad\quad\qquad+\frac{\rho_3}{2}\Vert\bTheta^{(t-1)}-\W+\U^{(t-1)}\Vert_F^2.
\end{align}
\end{small}
By setting the gradient with respect to $\W$ to zero,
\begin{small}
\begin{align}
    \rho_1\W+\rho_2\X^T\X\W-\rho_2\X^T\Y+\rho_3\W-\rho_3(\bTheta^{(t-1)}+\U^{(t-1)})=0,
\end{align}
\end{small}we arrive at 
\begin{align}
\label{solveW00}
\W^{(t)}=&\left[(\rho_1+\rho_3)\I_F+\rho_2 \X^T\X\right]^{-1}\nonumber\\&\left[\rho_3\left(\bTheta^{(t-1)}+\U^{(t-1)}\right)+\rho_2\X^T\Y\right].
\end{align}
In the next subsection, we will introduce a dynamic $\rho_3^{(t)}$ to achieve a predetermined pruning rate. Consequently, equation (\ref{solveW00}) will be modified to
\begin{align}
\label{solveW0}
\W^{(t)}=&[(\rho_1+\rho_3^{(t)})\I_F+\rho_2 \X^T\X]^{-1}\nonumber\\& \left[\rho_3^{(t)}\left(\bTheta^{(t-1)}+\U^{(t-1)}\right)+\rho_2\X^T\Y\right].
\end{align} 

\subsubsection{Updating $\bTheta$}
The updated $\bTheta$ is
\begin{align}
\bTheta^{(t)}&=\arg \min_{\bTheta}\Vert\bTheta'\Vert_{2,1}+\frac{\rho_3}{2}\Vert\bTheta-\W^{(t)}\Vert_F^2\nonumber\\
&\quad \quad\qquad+\text{tr}\left(\{\bLambda^{(t-1)}\}^T \circ \left(\bTheta-\W^{(t)}\right)\right)\nonumber\\
&=\arg \min_{\bTheta}\Vert\bTheta'\Vert_{2,1}+\frac{\rho_3}{2}\Vert\bTheta-\W^{(t)}+\U^{(t-1)}\Vert_F^2,
\end{align}
whose grouped components can be obtained using the soft-threshold operator  $\text{S}(\cdot)$ \cite{liao2014generalized} with the threshold $\frac{1}{\rho_3}$:
\begin{small}
\begin{align}
\label{solveTheta0}
\begin{split}
     \bTheta_{{\cb g}:}^{'(t)}&=\text{S}_{\frac{1}{\rho_3}}\left(\W_{{\cb g}}^{'(t)}-\U_{{\cb g}}^{'(t-1)}\right)\\
    &=\left(\W_{{\cb g}}^{'(t)}-\U_{{\cb g}}^{'(t-1)}\right)*\text{max}\left(1-\frac{\frac{1}{\rho_3}}{\Vert \W_{{\cb g}}^{'(t)}-\U_{{\cb g}}^{'(t-1)}\Vert_2},0\right),
\end{split}
\end{align}
\end{small} following the convention that $\frac{0}{0}=1$, {\cb where ${\cb g}=1,2,\cdots, G$}.


In equation (\ref{solveTheta0}), the parameter $\rho_3$ controls the penalty imposed on the magnitude of groups in $\Theta$. However, determining a suitable value for $\rho_3$ to achieve the desired pruning rate is a challenging task that often requires multiple attempts. To address this issue, motivated by \cite{liao2014generalized}, we adopt a dynamic threshold, $\rho_3^{(t)}$, to allow {\cb for predeterming the number of remaining} kernels $m$ ({\cb where $m<G$}).
Denoting $\widetilde{\W'}^{(t)}=\W'^{(t)}-\U'^{(t-1)}$, we define the dynamic threshold as
\begin{equation}
\label{rho1}
    \frac{1}{\rho_3^{(t)}}= \Vert {\W'}_{f_{m+1}:}^{(t)}-{\U'}_{f_{m+1}}^{(t-1)}\Vert_2=\Vert \widetilde{\W'}_{f_{m+1}:}^{(t)}\Vert_2,
\end{equation}
which satisfies
\begin{small}
\begin{equation}
\label{rho_order}
\Vert \widetilde{\W'}_{f_1:}^{(t)}\Vert_2\ge\cdots \ge \Vert \widetilde{\W'}_{f_m:}^{(t)}\Vert_2\ge\Vert \widetilde{\W'}_{f_{m+1}:}^{(t)}\Vert_2\ge\cdots \ge \Vert \widetilde{\W'}_{f_G:}^{(t)}\Vert_2,
\end{equation}
\end{small}
where $\left(f_{1}, {\cb \cdots,f_{m},f_{m+1}}, \cdots, f_{\cb{G}}\right)$ is a permutation of $(1,2,\cdots,\cb{G})$.
Using the dynamic threshold, the update for the grouped components of $\bTheta$ is
\begin{equation}
\label{solveTheta}
     \bTheta_{{\cb g}}^{'(t)}
    = \widetilde{\W'}_{{\cb g}}^{(t)}
*\text{max}\left(1-\frac{\Vert\widetilde{\W'}_{f_{m+1}:}^{(t)}\Vert_2}{\Vert \widetilde{\W'}_{{\cb g}}^{(t)}\Vert_2}, 0\right),
\end{equation}
for ${\cb g}=1,2,\cdots,{\cb G}$.


\subsubsection{Updating scale-form dual variable}
Dividing the both sides of the equation (\ref{dual}) by the step size $s$, we have
\begin{align}
\label{dualU0}
    \frac{\bLambda^{(t)}}{s}=\frac{\bLambda^{(t-1)}}{s}+ \left(\bTheta^{(t)}-\W^{(t)}\right).
\end{align}
Setting $s=\rho_3$, we arrive at
\begin{align}
\label{dualU1}
    \U^{(t)}=\U^{(t-1)}+\bTheta^{(t)}-\W^{(t)},
\end{align}
which is the updating equation for the scale-form dual variable matrix $\U$ defined in (\ref{defineU}).
Since the iteration steps (\ref{solveW0}) and (\ref{solveTheta}) involve $\U$ rather than $\bLambda$, we can directly update the scaled dual variable $\U$ during the iterations.

\subsubsection{Summary and Time Complexity }
By iteratively updating (\ref{solveW0}), (\ref{rho1}),  (\ref{solveTheta}), and (\ref{dualU1}),  we can successfully solve the random kernel pruning problem from a feature selection perspective. The dynamic threshold described in (\ref{rho1}) allows for arbitrary desired pruning rates according to the computational budget. 

The disadvantage of the ADMM-based algorithm is its high computational complexity, $\mathcal{O}(T{\cb H}^3)$  (assuming ${\cb H}>N$), where $T$ represents the number of iterations. As $\rho_3^{(t)}$ is dynamically varying, updating (\ref{solveW0}) requires the calculation of a large matrix inverse in each iteration. With ${\cb H}$ being 20K and 10K for ROCKET and MINIROCKET, respectively, it is time-consuming to prune redundant convolutional kernels.






\subsection{Accelerated Algorithm: {\cb POCKET}}
\label{accPOCKET}
To accelerate the ADMM-based algorithm, we further propose our core algorithm, {\cb POCKET}.
In contrast to the ADMM-based algorithm, {\cb POCKET} employs a two-stage strategy that separates group-level and element-wise regularization. {\cb As illustrated in Figure \ref{model}, t}he first stage implements group-wise regularization for pruning, and the optional second stage regularizes element-wise magnitudes to refit a high-accuracy classifier.

\subsubsection{Stage 1 (Group-wise Feature Selection)}
\label{stage1}
In the first stage, we only implement group-wise feature selection by solving
\begin{align}
\label{p3}
    &\min_{\W,\bTheta} \quad \Vert \bTheta' \Vert_{2,1} \nonumber\\
    \text{s.t.} \quad & \Vert\X\W-\Y\Vert_F< c_2^{(t)}, \quad \bTheta=\W.
\end{align}
Note that, different from problem (\ref{p2}), here we temporarily omit the constraint $\Vert \W\Vert_F<c_1$. Furthermore, we allow the penalty strength on $\Vert \X\W-\Y\Vert_F$ to dynamically vary in each iteration step, as indicated by the superscript $(t)$ on $c_2$. The problem can be solved using the ADMM-based algorithm proposed in the last subsection by simply replacing the update equation of $\W$ in (\ref{solveW0}) with
\begin{align}
\label{solveW01}
    \W^{(t)}=\big[\rho_3^{(t)}\I_F+\rho_2^{(t)} &\X^T\X\big]^{-1}\big[\rho_3^{(t)}\big(\bTheta^{(t)}+\U^{(t-1)}\big)\nonumber\\&+\rho_2^{(t-1)}\X^T\Y\big].
\end{align}
Here, the adaptive $\rho_3^{(t)}$ is the reciprocal of the soft threshold (\ref{rho1}), while $\rho_2^{(t)}$ is associated with the dynamically changing $c_2^{(t)}$.

However, the high computational complexity of (\ref{solveW01}) remains, since the computation of $[\rho_3^{(t)}\I_f+\rho_2^{(t)} \X^T\X]^{-1}$ is required in each iteration. To address the issue, we further introduce a requirement to maintain the relative invariance between $\rho_2^{(t)}$ and $\rho_3^{(t)}$:
\begin{equation}
    \frac{\rho_3^{(t)}}{\rho_2^{(t)}}=k,
\end{equation}
where $k>0$ is a user-defined constant.
In this scenario, equation (\ref{solveW01}) can be rewritten as 
\begin{equation}
\label{solveW1}
    \W^{(t)}=(k\I_F+\X^T\X)^{-1}\big[k\left(\bTheta^{(t-1)}+\U^{(t-1)}\right)+\X^T\Y\big].
\end{equation}
Here, the inverse term $(k\I_f+\X^T\X)^{-1}$ remains constant during iteration, thus it can be precomputed once and stored for reuse, leading to a significant reduction in time complexity.
Given the relatively large feature dimensions of ROCKET and MINIROCKET (20K and 10K, respectively), the Sherman-Morrison-Woodbury formula \cite{deng2013group, sherman1950adjustment} can be employed to further expedite the calculation for small-sample datasets ($N\ll {\cb H}$):
\begin{equation}
\label{SMW}
    (k\I_F+\X^T\X)^{-1}=\frac{1}{k}\left[\I_F-\X ^T(k\I_N+\X\X^T)^{-1}\X\right],
\end{equation}
which reduces the complexity from $\mathcal{O}({\cb H}^3)$ to $\mathcal{O}(N^3)$.

By repeating (\ref{solveW1}), (\ref{rho1}),  (\ref{solveTheta}) and (\ref{dualU1}), problem (\ref{p3}) can be solved to result in a group-sparse classifier that prunes redundant kernels via feature selection. The weight groups in $\W$ with zeros values and their corresponding convolution kernels can be safely deleted to reduce the overall size of the model. However, the weights in $\W$ that should be pruned often remain very small magnitudes. As $\bTheta$ in (\ref{solveTheta0}) possesses exactly zero-value groups and $\W$ should converge to $\bTheta$, we eliminate the groups in $\W$  that correspond to the zero-valued groups of $\bTheta$. Removing groups with very small values in $\W$ has a negligible impact on classification since the classifier fitted by $\Vert\X\W-\Y\Vert_F< c_2$ is noise-tolerant.



\subsubsection{Stage 2 (Optional, element-wise Regularization)}
In Stage 2, we implement the element-wise regularization and refit a ridge classifier using the features selected by Stage 1:
\begin{align}
\label{p4}
    \min_{\widehat{\W}} \quad \Vert\widehat{\X}\widehat{\W}-\Y\Vert_F \nonumber\\
    \text{s.t.} \quad   \Vert \widehat{\W} \Vert_F< c_1,
\end{align}
where $\widehat{\X}\in \bR^{N\times \widehat{{\cb H}}}$ signifies the feature matrix after pruning by Stage 1, $\widehat{\W}\in \bR^{\widehat{{\cb H}}\times C}$ represents the weight matrix, and $\hat{{\cb H}}\in \{1,2,\cdots,{\cb H}-1\}$ denotes the number of remaining features.
The solution of (\ref{p4}) can be written as
\begin{align}
\label{ridgeregression}
    \widehat{\W} = \left(\rho_1 \I+{\widehat{\X}}^T\widehat{\X}\right)^{-1}\widehat{\X}^T\Y,
\end{align}
where $\rho_1$ is a user-determined hyperparameter.

Since Stage 1 has already trained a classifier during feature selection, Stage 2 is optional and is primarily useful for addressing potential overfitting issues. In experiments, we observe that MINIROCKET is more sensitive to pruning as it is less redundant, and thus Stage 2 with element-wise regularization is more useful for MINIROCKET to significantly improve classification performance.


\subsubsection{Summary}
We summarize our core algorithm, {\cb POCKET}, in Algorithm \ref{alg2}. Different from the ADMM-based algorithm proposed in Section \ref{Solution using ADMM},  {\cb POCKET} divides the group-wise and element-wise regularizations into two sequential stages. The first stage performs kernel pruning via feature selection by imposing group sparsity in the classifier. The optional second stage uses ridge regularization to refit a classifier to alleviate potential overfitting. 

The advantage of {\cb POCKET} compared to the ADMM-based algorithm is its low time complexity. In Stage 1, by maintaining relative invariance between $\rho_2^{(t)}$ and $\rho_3^{(t)}$, {\cb POCKET} computes the inverse of the large matrix only once. Except for the two one-time inverse computations in (\ref{solveW1}) and (\ref{ridgeregression}), there are only matrix multiplies in iterations, which greatly reduces the time complexity.

\begin{algorithm}
\caption{{\cb POCKET}}\label{alg2}
\KwIn{\\Time series data matrix $\D$ with $N$ samples and label matrix $\Y$.\\Number of remained kernels  $m$, number of iteration $T$.\\Hyper-parameters $k$ and $\rho_1$. (In practice, they can be selected by cross-validation.)} 
\KwOut{\\The lightweight ROCKET/MINIROCKET model with $m$ kernels and a classifier $\widehat{\W}$.}
\BlankLine\BlankLine

// \textbf{Initialization:}\\Generate ${\cb G}$ random convolution kernels and extract the feature matrix $\X\in\bR^{N\times {\cb H}}$ from $\D$.\\Normalize $\X$ to make each group of features have a zero center and unit norm. \\Centralize each column of $\Y$ separately.
\BlankLine\BlankLine

// \textbf{Stage 1}: Pruning via Group Feature Selection\\
Precompute $(k\I_f+\X^T\X)^{-1}$. (Use (\ref{SMW}) if $N\ll {\cb H}$.)\\
$\bTheta \gets \textbf{0}$, $\U \gets \textbf{0}$. \\
\For{$t=1,2,\cdots,T$}{
  Compute $\W$ by (\ref{solveW1}).\\
  Update $\rho_3$ by (\ref{rho1}). \\
  Compute $\bTheta$ by (\ref{solveTheta}).\\
  Update $\U$ by (\ref{dualU1}).
}

Remain $m$ groups of weight in $\W$ indicated by nonzero components of $\bTheta$ and delete the others to get $\widehat{\W}$; delete the associated redundant random convolution kernels and features in $\X$ to get $\widehat{\X}$.
\BlankLine\BlankLine

// \textbf{Stage 2} (Optional): Element-wise Regularization\\
Recompute $\widehat{\W}$ by (\ref{ridgeregression}). 

\BlankLine
\textbf{return} $m$ remained random kernels, $\widehat{\W}$.
\end{algorithm}

\section{Experiments}
\label{Experiments}

\subsection{Data, models and General Settings}
\subsubsection{Datasets}
Following \cite{dempster2020rocket, salehinejad2022s}, we evaluate our algorithms on the UCR archive \cite{UCRArchive2018}, which contains 128 time series datasets, including 85 `bake off' datasets and 43 {\cb `extra'} ones. These datasets cover a wide range of application domains and serve as a comprehensive benchmark for evaluating time series classification algorithms. For missing values of data samples, we simply fill them with zeros.

\subsubsection{Target Models}
In line with S-ROCKET \cite{salehinejad2022s}, we prune MINIROCKET and ROCKET trained with only the PPV feature. However, to provide a more comprehensive evaluation, we also prune the original full ROCKET model trained with both PPV and MAX features. To differentiate between the full ROCKET model with both MAX and PPV features, and the one with only PPV, we will refer to the former as `\textbf{ROCKET}' or `\textbf{ROCKET-PPV-MAX}', and the latter as `\textbf{ROCKET-PPV}'.

\subsubsection{General Settings}
\label{General Settings}
Unless explicitly stated, all experiments are conducted using Python 3.6.12 and scikit-learn 0.24.2 \cite{scikit-learn} on a {\cb Ubuntu 18.04.6} computer with two Intel Xeon E5-2690 CPUs and 56 threads in total. For a fair comparison, we use the official code and settings to reproduce ROCKET, MINIROCKET, and S-ROCKET. 

For our ADMM-based algorithm and {\cb POCKET}, the number of iterations for both cross-validation and fitting is  $T=50$.
Hyper-parameters $\rho_1$ and $\rho_2$ for the ADMM-based algorithm, and $k$ for Stage 1 of {\cb POCKET}, are determined through 5-fold cross-validations.  Following ROCKET \cite{salehinejad2022s} and MINIROCKET \cite{dempster2021minirocket}, $\rho_1$ in Stage 2 of {\cb POCKET} is decided by the efficient Leave-One-Out cross-validation on the log scale range of $[-3,3]$.  Each evaluation is repeated 10 times to report the average results. More implementation details can be found in our publicly available code\footnote{\href{https://github.com/ShaowuChen/POCKET}{https://github.com/ShaowuChen/POCKET}}.

\subsection{ADMM-based algorithm vs. {\cb POCKET}}
\subsubsection{Settings}
We first compare our ADMM-based algorithm with the accelerated {\cb POCKET} on pruning 50\% of kernels from ROCKET-PPV-MAX for the first 10 datasets of the UCR archive.
 To choose the optimal hyperparameters, which encompass $\rho_1$ and $\rho_2$ for the ADMM-based algorithm, and $k$ for Stage 1 of {\cb POCKET}, we use a 5-fold cross-validation process within the range of $[0.01, 0.1, 1, 10, 100]$, employing 25 threads for computational efficiency.

\subsubsection{Results and Discussions}
Table \ref{ADMVSPROCKET} compares the test accuracy between the ADMM-based algorithm and {\cb POCKET}. Notably, both algorithms can successfully remove redundancy without causing severe degradation in most cases. {\cb POCKET} shows superior performance than the ADMM-based algorithm and even outperforms the original full model when 50\% of kernels are pruned. We also compare the training time efficiency in Table \ref{ADMVSPROCKETtime}, in which {\cb POCKET} is approximately 10 times faster than the ADMM-based algorithm.

{\cb The `Beef' dataset is an outlier, in which the ADMM-based algorithm performs more than 30\% worse than POCKET, while they have similar accuracy on the other datasets. To explore the anomaly, we examine the validation test accuracy for both algorithms, which is consistently \textbf{0\%} for various parameter values. This is due to the 'Beef' dataset's limited size, with only 30 training samples spread across 5 categories, resulting in each fold of the 5-fold cross-validation process containing only 1 training sample. In this scenario, where the values for hyperparameters---$\rho_1$ and $\rho_2$ for the ADMM-base algorithm, and $k$ for POCKET---are chosen almost at random, the robustness of algorithms with respect to hyperparameters becomes a crucial factor.

We further investigate the training accuracy of both algorithms with different hyperparameters in Table \ref{ADM_hyper} and \ref{PROCKET_hyper}.
It is shown that the validation training accuracy of the ADMM-based algorithm varies dramatically from 16.67\% to 87.50\% when $\rho_1$ and $\rho_2$ are changed, whereas those of POCKET remains consistently at 100\% when $k$ varies within the same range. This suggests that the ADMM-based algorithm is more sensitive to hyperparameters than POCKET, making it challenging to select appropriate values, especially when two hyperparameters ($\rho_1$ and $\rho_2$) are involved simultaneously.

For further verification, we also rerun the ADMM-based algorithm on the `Beef' dataset, setting $\rho_1$ and $\rho_2$ to the values that produced the highest validation training accuracy in Table \ref{ADM_hyper}, namely $0.01$ and $1$, respectively. Remarkably, this adjustment results in a significant 30\% improvement in test accuracy, indicating that the ADMM-based algorithm relies on more careful tuning of hyperparameters compared to {\cb POCKET}. }


As the ADMM-based algorithm is time-consuming and the core {\cb POCKET exhibits} superior performance, the subsequent experiments will omit the former and concentrate solely on evaluating {\cb POCKET}.

\begin{table}[htbp]
\scriptsize
\caption{Testing accuracy comparison on pruning 50\% kernels of ROCKET-PPV-MAX. The `\textbf{Ave. w/o Beef}' row presents the average testing accuracy across datasets except for the `Beef'.}
\setlength\tabcolsep{3pt}
\centering
\begin{tabular}{lcccc}
 \toprule
\multirow{2}{*}{Dataset} &  \multirow{2}{*}{\tabincell{c}{Original\\ROCKET\\ (\%)}}&  \multirow{2}{*}{\tabincell{c}{\textbf{ADMM-based}\\pruning\\ (\%)}} & \multicolumn{2}{c}{\textbf{{\cb POCKET} (\%)  }}\\ 

\cmidrule(r){4-5} 

&&&Stage 1& Stage 2\\

\midrule
Adiac                 & 78.44$\pm$0.61  & 81.18$\pm$0.86  & 80.84$\pm$0.98  & 80.23$\pm$0.67  \\
ArrowHead             & 81.20$\pm$0.90  & 81.89$\pm$1.20  & 81.54$\pm$1.58  & 80.86$\pm$0.82  \\
\underline{Beef}                  & 83.33$\pm$1.49  & \underline{51.00$\pm$2.60}  & 83.67$\pm$1.00  & 84.33$\pm$1.53  \\
BeetleFly             & 90.00$\pm$0.00  & 90.00$\pm$0.00  & 90.00$\pm$0.00  & 90.00$\pm$0.00  \\
BirdChicken           & 90.00$\pm$0.00  & 90.00$\pm$0.00  & 90.00$\pm$0.00  & 90.00$\pm$0.00  \\
Car                   & 86.00$\pm$3.35  & 91.5$\pm$1.38   & 89.83$\pm$4.74  & 90.50$\pm$1.98  \\
CBF                   & 100.00$\pm$0.00 & 99.87$\pm$0.10  & 99.92$\pm$0.05  & 99.99$\pm$0.03  \\
CinCECGT          & 83.36$\pm$0.26  & 90.17$\pm$1.05  & 90.51$\pm$0.58  & 85.22$\pm$0.33  \\
ChlCon & 81.43$\pm$0.80  & 76.15$\pm$0.73  & 79.80$\pm$0.68  & 81.08$\pm$0.49  \\
Coffee                & 100.00$\pm$0.00 & 100.00$\pm$0.00 & 100.00$\pm$0.00 & 100.00$\pm$0.00 \\
\midrule
\textbf{Average} & 87.38$\pm$0.74  & 85.18$\pm$0.79  &\textbf{88.61$\pm$0.96}  & \textbf{88.22$\pm$0.59}  \\
\textbf{Ave. w/o Beef}                       & 87.83$\pm$0.66  & \textbf{88.97$\pm$0.59}  & \textbf{89.16$\pm$0.96} & \textbf{88.65$\pm$0.48} \\
\bottomrule
\end{tabular}
\label{ADMVSPROCKET}
\end{table}

\begin{table}[]
\scriptsize
\caption{Training time comparison on pruning 50\% of kernels of ROCKET-PPV-MAX. `CV' shows the cross-validation time for finding optimal hyper-parameters, while `Refit' shows the time cost for refitting a classifier with the decided hyper-parameters.}
\setlength\tabcolsep{2pt}
\begin{tabular}{lccccccc}
\toprule
\multirow{3}{*}{Dataset} &  \multicolumn{3}{c}{\textbf{ADMM-based} (in Seconds)} & \multicolumn{4}{c}{\textbf{{\cb POCKET}} (in Seconds)}\\ 
\cmidrule(r){2-4} \cmidrule(r){5-8} 
& CV & Refit & \textit{Sum}& \tabincell{c}{Stage 1\\CV} & \tabincell{c}{Stage 1\\Refit} & \tabincell{c}{Stage\\2} & \textit{Sum}\\
\midrule
Adiac       & 3306.55 & 199.65 & \textit{3506.20} & 289.98 & 42.30 & 0.22 & \textit{332.50} \\
ArrowHead   & 2357.46 & 149.73 & \textit{2507.19} & 252.06 & 20.01 & 0.02 & \textit{272.09} \\
Beef        & 2547.72 & 148.37 & \textit{2696.09} & 310.08 & 19.92 & 0.02 & \textit{330.02} \\
BeetleFly   & 2451.91 & 148.23 & \textit{2600.14} & 279.93 & 19.27 & 0.01 & \textit{299.21} \\
BirdChicken & 2216.72 & 149.12 & \textit{2365.84} & 249.44 & 18.73 & 0.01 & \textit{268.18} \\
Car         & 2650.60 & 152.21 & \textit{2802.81} & 307.73 & 20.52 & 0.04 & \textit{328.29} \\
CBF         & 2295.02 & 153.04 & \textit{2448.06} & 241.28 & 19.52 & 0.01 & \textit{260.81} \\
CinCECGT    & 2471.25 & 149.79 & \textit{2621.04} & 293.84 & 19.82 & 0.02 & \textit{313.68} \\
ChlCon      & 3369.88 & 189.48 & \textit{3559.36} & 248.16 & 20.38 & 0.31 & \textit{268.85} \\
Coffee      & 2411.54 & 152.27 & \textit{2563.81} & 269.69 & 18.96 & 0.01 & \textit{288.66} \\
\midrule
\textbf{Average}     & 2629.68 & 159.96 & \textit{2789.64} & \textbf{274.72} & \textbf{22.27} & 0.07 & \textbf{\textit{297.07}}\\
\bottomrule
\end{tabular}
\label{ADMVSPROCKETtime}
\end{table}

\begin{table}[htbp]
\setlength\tabcolsep{5pt}

\caption{Validation training accuracy of our ADMM-based Algorithm on pruning ROCKET-PPV-MAX for the `Beef' dataset. Here `VT. Acc.' denotes `Validation Training Accuracy'.}
\centering
\begin{tabular}{cccccc}
\toprule
\diagbox{$\rho_2$}{VT. Acc. (\%)}{$\rho_1$} & 0.01 & 0.1 & 1 & 10 & 100 \\
\midrule
0.01 & 67.50 & 67.50 & 67.50 & 67.50 & 67.50\\ 
0.1  &  67.50 & 67.50 & 67.50 & 67.50 & 67.50\\
1    & \textbf{87.50} & \textbf{87.50} & \textbf{87.50} & \textbf{87.50} & 86.67\\  
10   & 16.67 & 16.67 & 16.67 & 16.67 & 33.33 \\   
100  & 16.67 & 16.67 & 16.67 & 16.67 & 25.00 \\
\bottomrule
\end{tabular}
\label{ADM_hyper}
\end{table}

\begin{table}[htbp]
\setlength\tabcolsep{5pt}

\caption{Validation training accuracy of our {\cb POCKET} on pruning ROCKET-PPV-MAX for the `Beef' dataset.}
\centering
\begin{tabular}{cccccc}
\toprule
$k$ & 0.01 & 0.1 & 1 & 10 & 100 \\
\midrule
VT. Acc. (\%) &100.00 &100.00 &100.00 &100.00 &100.00 \\
\bottomrule
\end{tabular}
\label{PROCKET_hyper}
\end{table}

\begin{table*}[]
\scriptsize
\caption{ Performance comparison {\cb (\%)} on pruning ROCKET-PPV-MAX.}
\centering
\begin{tabular}{lcccccc}
 \toprule
\multirow{4}{*}{Dataset} &  \multirow{4}{*}{\tabincell{c}{Original\\ROCKET {\cb \cite{dempster2020rocket}}\\Acc.$\pm$Std.}}&\multicolumn{2}{c}{S-ROCKET \cite{salehinejad2022s}}  & \multicolumn{3}{c}{Our \textbf{{\cb POCKET} }}\\ 

\cmidrule(r){3-4} \cmidrule(r){5-7}

& &\tabincell{c}{Remain\\Rate} &\tabincell{c}{Acc.$\pm$Std.} &\tabincell{c}{Remain\\Rate} & \tabincell{c}{Stage 1\\Acc.$\pm$Std.} & \tabincell{c}{Stage 2\\Acc.$\pm$Std.}\\

\midrule

Adiac                        & 78.13$\pm$0.43  & 100.00 & 78.13$\pm$0.43  & \textbf{50.00} & 80.18$\pm$0.54  & 79.87$\pm$0.50  \\
ArrowHead                    & 81.37$\pm$1.03  & 24.47  & 81.77$\pm$1.31  & 24.47 & 80.86$\pm$1.93  & 81.83$\pm$1.59  \\
Beef                         & 82.00$\pm$3.71  & 19.46  & 81.00$\pm$2.60  & 19.46 & 82.67$\pm$2.49  & 83.33$\pm$3.65  \\
BeetleFly                    & 90.00$\pm$0.00  & 21.08  & 90.00$\pm$0.00  & 21.08 & 89.50$\pm$1.50  & 90.00$\pm$0.00  \\
BirdChicken                  & 90.00$\pm$0.00  & 24.31  & 89.00$\pm$3.00  & 24.31 & 90.00$\pm$0.00  & 90.00$\pm$0.00  \\
Car                          & 88.33$\pm$1.83  & 34.29  & 88.33$\pm$2.69  & 34.29 & 92.50$\pm$0.83  & 91.67$\pm$1.29  \\
CBF                          & 100.00$\pm$0.00 & 17.98  & 99.96$\pm$0.05  & 17.98 & 99.92$\pm$0.07  & 99.98$\pm$0.04  \\
CinCECGT                 & 83.61$\pm$0.55  & 24.44  & 82.79$\pm$0.74  & 24.44 & 90.86$\pm$2.96  & 88.23$\pm$1.64  \\
ChlCon        & 81.50$\pm$0.49  & 40.67  & 79.26$\pm$1.46  & 40.67 & 79.43$\pm$0.59  & 80.71$\pm$0.60  \\
Coffee                       & 100.00$\pm$0.00 & 18.06  & 100.00$\pm$0.00 & 18.06 & 100.00$\pm$0.00 & 100.00$\pm$0.00 \\
Computers                    & 76.32$\pm$0.84  & 26.75  & 76.80$\pm$0.95  & 26.75 & 74.60$\pm$1.82  & 77.20$\pm$0.76  \\
CricketX                     & 81.92$\pm$0.49  & 72.95  & 82.05$\pm$0.62  & 72.95 & 82.10$\pm$0.66  & 82.18$\pm$0.80  \\
CricketY                     & 85.38$\pm$0.60  & 57.50  & 85.08$\pm$0.56  & 52.50 & 83.87$\pm$0.75  & 84.90$\pm$0.71  \\
CricketZ                     & 85.44$\pm$0.64  & 70.40  & 85.03$\pm$0.69  & 70.40 & 83.87$\pm$0.71  & 85.10$\pm$0.71  \\
DiaSizRed          & 97.09$\pm$0.61  & 24.23  & 96.50$\pm$0.84  & 24.23 & 95.59$\pm$2.37  & 97.84$\pm$0.55  \\
DisPhaOAG & 75.68$\pm$0.63  & 32.85  & 74.89$\pm$0.99  & 32.85 & 74.89$\pm$0.68  & 73.74$\pm$0.98  \\
DisPhaOutCor  & 76.74$\pm$0.88  & 32.71  & 77.03$\pm$1.30  & 32.71 & 77.54$\pm$0.96  & 76.27$\pm$1.61  \\
DoLoDay                & 60.50$\pm$1.70  & 55.20  & 60.25$\pm$1.84  & 50.20 & 60.88$\pm$1.68  & 60.62$\pm$2.11  \\
DoLoGam               & 86.09$\pm$0.54  & 18.43  & 86.45$\pm$0.80  & 18.43 & 88.33$\pm$0.68  & 88.26$\pm$0.63  \\
DoLoWKE            & 97.68$\pm$0.29  & 18.28  & 97.54$\pm$0.35  & 18.28 & 98.55$\pm$0.00  & 98.48$\pm$0.22  \\
Earthquakes                  & 74.82$\pm$0.00  & 32.64  & 74.82$\pm$0.00  & 32.64 & 74.96$\pm$0.29  & 74.96$\pm$0.54  \\
ECG200                       & 90.40$\pm$0.49  & 10.88  & 89.90$\pm$0.70  & 10.88 & 91.20$\pm$1.17  & 90.60$\pm$0.66  \\
ECG5000                      & 94.75$\pm$0.05  & 33.91  & 94.68$\pm$0.09  & 33.91 & 93.55$\pm$0.60  & 94.78$\pm$0.06  \\
ECGFiveDays                  & 100.00$\pm$0.00 & 22.19  & 100.00$\pm$0.00 & 22.19 & 100.00$\pm$0.00 & 100.00$\pm$0.00 \\
EOGHSignal          & 58.26$\pm$1.05  & 69.33  & 58.01$\pm$1.19  & 64.33 & 59.17$\pm$1.46  & 58.45$\pm$1.18  \\
EOGVSignal            & 54.70$\pm$0.58  & 51.56  & 55.03$\pm$0.91  & 51.56 & 55.28$\pm$0.94  & 54.81$\pm$0.58  \\
FaceAll                      & 94.68$\pm$0.40  & 46.61  & 94.64$\pm$0.32  & 46.61 & 94.16$\pm$0.67  & 94.14$\pm$0.78  \\
FaceFour                     & 97.61$\pm$0.34  & 18.30  & 97.61$\pm$0.34  & 18.30 & 98.41$\pm$0.56  & 97.84$\pm$0.34  \\
FacesUCR                     & 96.20$\pm$0.09  & 48.43  & 96.20$\pm$0.08  & 48.43 & 96.14$\pm$0.16  & 96.34$\pm$0.08  \\
FiftyWords                   & 82.99$\pm$0.41  & 100.00 & 82.99$\pm$0.41  & 50.00 & 82.00$\pm$0.47  & 82.46$\pm$0.47  \\
\midrule
\textit{\textbf{AVERAGE} }                     & 84.74$\pm$0.62  & 38.93  & 84.52$\pm$0.84  & \textbf{35.10} & \textbf{85.03$\pm$0.92}  & \textbf{85.15$\pm$0.77} \\
\bottomrule
\end{tabular}
\label{ROCKET_result}
\end{table*}

\begin{table*}[]
\scriptsize
\caption{Comparison of {\cb average} training time {\cb (for one run)} in different steps or stages for ROCKET-PPV-MAX. 'Stage 1 CV' shows the cross-validation time for finding optimal hyper-parameter $k$ in Stage 1, while 'Stage 1 Refit' shows the time cost for refitting the classifier with the decided $k$. {\cb `\textit{Sum}' shows the total training time.}}
\centering
\begin{tabular}{lccccccccc}
 \toprule

\multirow{3}{*}{Dataset} &\multicolumn{4}{c}{S-ROCKET \cite{salehinejad2022s}  (in Seconds)} & &\multicolumn{4}{c}{\textbf{{\cb POCKET}} (in Seconds)}  \\
\cmidrule(r){2-5} \cmidrule(r){7-10}
& \multirow{2}{*}{Pre-train} & \multirow{2}{*}{\tabincell{c}{Kernel\\Selection}} & \multirow{2}{*}{\tabincell{c}{Post\\training}} & \multirow{2}{*}{\textit{Sum}}& & \multirow{2}{*}{\tabincell{c}{Stage 1\\CV}} & \multirow{2}{*}{\tabincell{c}{Stage 1\\Refit}} & \multirow{2}{*}{\tabincell{c}{Stage\\2}} & \multirow{2}{*}{\textit{Sum}}\\
\\
\midrule



Adiac            & 0.39 & 5550.66 & 0.39 & \textit{5551.44} &  & 226.68 & 43.79 & 0.22 & \textit{270.69}          \\
ArrowHead        & 0.02 & 398.64  & 0.02 & \textit{398.68}  &  & 117.28 & 19.03 & 0.02 & \textit{136.33}          \\
Beef             & 0.02 & 411.45  & 0.02 & \textit{411.49}  &  & 114.50 & 19.39 & 0.01 & \textit{133.90}          \\
BeetleFly        & 0.01 & 200.08  & 0.01 & \textit{200.10}  &  & 102.49 & 18.02 & 0.01 & \textit{120.52}          \\
BirdChicken      & 0.01 & 199.94  & 0.01 & \textit{199.96}  &  & 103.79 & 18.01 & 0.01 & \textit{121.81}          \\
Car              & 0.04 & 699.04  & 0.06 & \textit{699.14}  &  & 107.39 & 19.62 & 0.05 & \textit{127.06}          \\
CBF              & 0.02 & 344.29  & 0.02 & \textit{344.33}  &  & 118.49 & 18.90 & 0.01 & \textit{137.40}          \\
CinCECGT         & 0.02 & 482.37  & 0.03 & \textit{482.42}  &  & 113.93 & 19.14 & 0.01 & \textit{133.08}          \\
ChlCon           & 0.44 & 2228.06 & 0.37 & \textit{2228.87} &  & 141.75 & 20.92 & 0.24 & \textit{162.91}          \\
Coffee           & 0.02 & 269.04  & 0.02 & \textit{269.08}  &  & 106.25 & 18.61 & 0.01 & \textit{124.87}          \\
Computers        & 0.20 & 1808.80 & 0.18 & \textit{1809.18} &  & 115.01 & 18.88 & 0.09 & \textit{133.98}          \\
CricketX         & 0.37 & 2821.16 & 0.34 & \textit{2821.87} &  & 141.94 & 24.92 & 0.31 & \textit{167.17}          \\
CricketY         & 0.37 & 2834.85 & 0.33 & \textit{2835.55} &  & 140.86 & 24.94 & 0.25 & \textit{166.05}          \\
CricketZ         & 0.36 & 2836.02 & 0.32 & \textit{2836.70} &  & 143.78 & 25.30 & 0.28 & \textit{169.36}          \\
DiaSizRed        & 0.02 & 209.85  & 0.01 & \textit{209.88}  &  & 105.53 & 19.35 & 0.02 & \textit{124.90}          \\
DisPhaOAG        & 0.37 & 1898.41 & 0.34 & \textit{1899.12} &  & 126.13 & 20.26 & 0.16 & \textit{146.55}          \\
DisPhaOutCor     & 0.58 & 2498.19 & 0.50 & \textit{2499.27} &  & 126.50 & 19.78 & 0.29 & \textit{146.57}          \\
DoLoDay          & 0.04 & 512.63  & 0.07 & \textit{512.74}  &  & 128.89 & 25.26 & 0.07 & \textit{154.22}          \\
DoLoGam          & 0.01 & 199.82  & 0.02 & \textit{199.85}  &  & 105.36 & 18.12 & 0.02 & \textit{123.50}          \\
DoLoWKE          & 0.01 & 199.69  & 0.02 & \textit{199.72}  &  & 104.61 & 17.20 & 0.01 & \textit{121.82}          \\
Earthquakes      & 0.29 & 2309.31 & 0.26 & \textit{2309.86} &  & 115.42 & 19.46 & 0.14 & \textit{135.02}          \\
ECG200           & 0.06 & 784.47  & 0.08 & \textit{784.61}  &  & 116.22 & 19.54 & 0.05 & \textit{135.81}          \\
ECG5000          & 0.48 & 2686.28 & 0.42 & \textit{2687.18} &  & 152.17 & 20.57 & 0.26 & \textit{173.00}          \\
ECGFiveDays      & 0.01 & 205.93  & 0.02 & \textit{205.96}  &  & 106.90 & 17.91 & 0.01 & \textit{124.82}          \\
EOGHSignal       & 0.35 & 2626.89 & 0.29 & \textit{2627.53} &  & 143.25 & 25.34 & 0.24 & \textit{168.83}          \\
EOGVSignal       & 0.34 & 2634.98 & 0.29 & \textit{2635.61} &  & 139.29 & 25.59 & 0.20 & \textit{165.08}          \\
FaceAll          & 0.53 & 4372.66 & 0.46 & \textit{4373.65} &  & 164.16 & 31.94 & 0.37 & \textit{196.47}          \\
FaceFour         & 0.01 & 321.44  & 0.02 & \textit{321.47}  &  & 106.71 & 19.10 & 0.01 & \textit{125.82}          \\
FacesUCR         & 0.12 & 1599.16 & 0.12 & \textit{1599.40} &  & 151.37 & 31.75 & 0.08 & \textit{183.20}          \\
FiftyWords       & 0.44 & 8067.68 & 0.41 & \textit{8068.53} &  & 264.88 & 43.44 & 0.30 & \textit{308.62}          \\
\midrule
\textit{\textbf{AVERAGE} } & 0.20 & 1740.39 & 0.18 & \textit{1740.77} &  & 131.72 & 22.80 & 0.13 & \textit{\textbf{154.65}} \\
\bottomrule
\end{tabular}
\label{Time}
\end{table*}

\begin{table*}[]
\scriptsize
\caption{Performance comparison {\cb (\%)} on pruning ROCKET-PPV. The data columns {\cb marked with '*' are taken from S-ROCKET \cite{salehinejad2022s},} while those without it are achieved by reproduction under the same settings.}
\centering
\begin{tabular}{l cccccc ccc}
 \toprule

\multirow{5}{*}{Dataset} &  \multirow{5}{*}{\tabincell{c}{Original\\ROCKET {\cb \cite{dempster2020rocket}}\\Acc.*}}  & \multirow{5}{*}{\tabincell{c}{Remain\\Rate*}}  & \multirow{5}{*}{\tabincell{c}{Random\\Acc.*}} & \multirow{5}{*}{\tabincell{c}{$\ell_1$-norm\\ \cite{0022KDSG17}\\Acc.*}}  & \multirow{5}{*}{\tabincell{c}{Soft Filter\\\cite{HeKDFY18}\\Acc.*}}   & \multirow{5}{*}{\tabincell{c}{S-ROCKET\\\cite{salehinejad2022s}\\Acc.*} }   &  \multicolumn{3}{c}{Our \textbf{{\cb POCKET} }(\%)}\\ 

\cmidrule(r){8-10}

& &&&&&   &\tabincell{c}{Original\\ROCKET  {\cb \cite{dempster2020rocket}}\\Acc.$\pm$Std.} & \tabincell{c}{Stage 1\\Acc.$\pm$Std.} & \tabincell{c}{Stage 2\\Acc.$\pm$Std.}\\
\midrule

Adiac                             & 78  & 41 & 69 & 72 & 71 & 78  & 78.34$\pm$0.64  & 80.20$\pm$0.57  & 79.67$\pm$0.50  \\
ArrowHead                         & 83  & 41 & 79 & 82 & 82 & 83  & 83.43$\pm$2.09  & 85.03$\pm$1.30  & 83.49$\pm$1.24  \\
Beef                              & 82  & 16 & 70 & 80 & 80 & 82  & 82.33$\pm$1.53  & 82.33$\pm$2.13  & 82.67$\pm$1.33  \\
BeetleFly                         & 95  & 27 & 86 & 91 & 91 & 95  & 95.00$\pm$0.00  & 96.00$\pm$2.00  & 95.00$\pm$0.00  \\
BirdChicken                       & 90  & 20 & 78 & 83 & 83 & 88  & 90.00$\pm$0.00  & 90.00$\pm$0.00  & 90.00$\pm$0.00  \\
Car                               & 90  & 19 & 81 & 85 & 84 & 87  & 88.00$\pm$5.15  & 92.33$\pm$1.33  & 93.17$\pm$1.38  \\
CBF                               & 100 & 19 & 85 & 97 & 99 & 100 & 99.89$\pm$0.00  & 99.89$\pm$0.00  & 99.89$\pm$0.00  \\
CinCECGT                          & 81  & 21 & 71 & 80 & 81 & 80  & 80.48$\pm$0.88  & 83.17$\pm$1.93  & 82.51$\pm$1.20  \\
ChlCon                            & 76  & 31 & 62 & 73 & 73 & 73  & 76.26$\pm$0.54  & 73.82$\pm$1.21  & 75.55$\pm$0.65  \\
Coffee                            & 100 & 58 & 95 & 98 & 98 & 99  & 100.00$\pm$0.00 & 100.00$\pm$0.00 & 100.00$\pm$0.00 \\
Computers                         & 77  & 29 & 69 & 76 & 74 & 77  & 77.32$\pm$0.51  & 78.36$\pm$0.63  & 77.96$\pm$0.85  \\
CricketX                          & 83  & 72 & 81 & 82 & 82 & 83  & 82.82$\pm$0.79  & 82.46$\pm$0.74  & 82.95$\pm$0.79  \\
CricketY                          & 85  & 71 & 78 & 80 & 80 & 85  & 84.74$\pm$0.80  & 84.10$\pm$1.12  & 84.79$\pm$0.90  \\
CricketZ                          & 85  & 70  & 79 & 81 & 82 & 85  & 84.87$\pm$0.57  & 84.38$\pm$0.79  & 84.79$\pm$0.59  \\
DiaSizRed                         & 95  & 29 & 88 & 94 & 94 & 95  & 95.36$\pm$0.70  & 96.34$\pm$0.91  & 95.52$\pm$1.15  \\
DisPhaOAG                         & 75  & 35 & 69 & 73 & 73 & 75  & 74.89$\pm$0.88  & 74.24$\pm$1.36  & 74.82$\pm$1.47  \\
DisPhaOutCor                      & 77  & 35 & 69 & 74 & 75 & 77  & 76.96$\pm$1.08  & 76.96$\pm$1.39  & 77.39$\pm$1.60  \\
DoLoDay                           & 65  & 69 & 58 & 54 & 56 & 64  & 61.75$\pm$1.87  & 62.12$\pm$1.77  & 62.12$\pm$2.17  \\
DoLoGam                           & 80   & 23 & 75 & 78 & 80 & 81  & 84.42$\pm$0.87  & 82.83$\pm$0.80  & 83.91$\pm$0.78  \\
DoLoWKE                           & 97  & 1 & 82 & 92 & 92 & 94  & 97.83$\pm$0.00  & 98.12$\pm$0.35  & 98.12$\pm$0.35  \\
Earthquakes                       & 75  & 1 & 69 & 75 & 75 & 75  & 74.82$\pm$0.00  & 75.47$\pm$1.04  & 75.32$\pm$0.97  \\
ECG200                            & 90  & 18 & 81 & 89 & 89 & 88  & 89.80$\pm$0.75  & 88.80$\pm$1.60  & 89.80$\pm$0.75  \\
ECG5000                           & 95  & 29 & 73 & 89 & 89 & 95  & 94.67$\pm$0.05  & 94.56$\pm$0.06  & 94.63$\pm$0.05  \\
ECGFiveDays                       & 100 & 21 & 89 & 98 & 98 & 100 & 100.00$\pm$0.00 & 100.00$\pm$0.00 & 100.00$\pm$0.00 \\
EOGHSignal                        & 57  & 56 & 49 & 55 & 55 & 55  & 57.07$\pm$0.98  & 55.58$\pm$1.31  & 56.71$\pm$0.95  \\
EOGVSignal                        & 44  & 59 & 41 & 41 & 42 & 44  & 44.97$\pm$1.13  & 44.39$\pm$1.70  & 44.78$\pm$1.34  \\
FaceAll                           & 79  & 52 & 69 & 78 & 78 & 80  & 79.46$\pm$0.36  & 79.59$\pm$0.37  & 79.50$\pm$0.41  \\
FaceFour                          & 100 & 59 & 82 & 98 & 99 & 100 & 100.00$\pm$0.00 & 100.00$\pm$0.00 & 100.00$\pm$0.00 \\
FacesUCR                          & 96  & 50  & 88 & 92 & 91 & 96  & 96.30$\pm$0.13  & 96.10$\pm$0.23  & 96.30$\pm$0.15  \\
FiftyWords                        & 85  & 78 & 76 & 66 & 65 & 85  & 84.79$\pm$0.35  & 85.10$\pm$0.55  & 84.95$\pm$0.46  \\
\midrule
\textit{\textbf{AVERAGE} } & 84  & 39 & 75 & 80 & 80  & 83  & 83.89$\pm$0.76  & \textbf{84.08$\pm$0.91}  & \textbf{84.21$\pm$0.73} \\
\bottomrule
\end{tabular}
\label{ROCKET_PPV_result}
\end{table*}

\begin{table*}[]
\scriptsize
\centering
\caption{Performance comparison {\cb (\%)} on pruning MINIROCKET. {\cb The columns marked with `*' are taken from S-ROCKET \cite{salehinejad2022s}}.}
\begin{tabular}{lclccccccc}
\toprule

\multirow{5}{*}{Dataset} &  \multirow{5}{*}{\tabincell{c}{Original\\MINIROCKET \\Acc.  {\cb \cite{dempster2021minirocket}}*}}  & \multirow{5}{*}{\tabincell{c}{Remain\\Rate* }} & \multirow{5}{*}{\tabincell{c}{Random\\Acc.*}}  & \multirow{5}{*}{\tabincell{c}{$\ell_1$-norm\\\cite{0022KDSG17}\\Acc.*}}  & \multirow{5}{*}{\tabincell{c}{Soft Filter\\\cite{HeKDFY18}\\Acc.*}}    & \multirow{5}{*}{\tabincell{c}{S-ROCKET\\\cite{salehinejad2022s}\\Acc.*} }   &  \multicolumn{3}{c}{Our \textbf{{\cb POCKET} } }\\ 

\cmidrule(r){8-10}

& &&&&&   &\tabincell{c}{Original\\MINIROCKET  {\cb \cite{dempster2021minirocket}}\\Acc.$\pm$Std.} & \tabincell{c}{Stage 1\\Acc.$\pm$Std.} & \tabincell{c}{Stage 2\\Acc.$\pm$Std.}\\
\midrule

Adiac        & 82  & 10 & 63 & 71 & 72 & 78  & 82.12$\pm$0.53  & 81.10$\pm$0.82  & 81.61$\pm$0.90  \\
ArrowHead    & 87  & 35 & 81 & 83 & 83 & 86  & 86.51$\pm$0.64  & 88.74$\pm$1.14  & 87.20$\pm$0.89  \\
Beef         & 87  & 10 & 73 & 79 & 81 & 83  & 86.67$\pm$0.00  & 84.00$\pm$3.59  & 86.67$\pm$1.49  \\
BeetleFly    & 88  & 30 & 79 & 86 & 86 & 87  & 89.00$\pm$2.00  & 85.50$\pm$1.50  & 89.50$\pm$1.50  \\
BirdChicken  & 90  & 19 & 81 & 84 & 86 & 89  & 90.00$\pm$0.00  & 90.00$\pm$0.00  & 90.00$\pm$0.00  \\
Car          & 92  & 49 & 78 & 90 & 91 & 92  & 91.67$\pm$0.00  & 91.67$\pm$0.00  & 91.67$\pm$0.00  \\
CBF          & 100 & 1  & 89 & 96 & 98 & 100 & 99.89$\pm$0.00  & 98.20$\pm$0.79  & 99.67$\pm$0.26  \\
CinCECGT     & 87  & 39 & 65 & 75 & 76 & 84  & 86.33$\pm$0.32  & 86.64$\pm$0.94  & 86.83$\pm$0.52  \\
ChlCon       & 76  & 30 & 53 & 72 & 72 & 72  & 76.02$\pm$0.49  & 73.90$\pm$0.60  & 75.39$\pm$0.36  \\
Coffee       & 100 & 33 & 87 & 99 & 99 & 100 & 100.00$\pm$0.00 & 100.00$\pm$0.00 & 100.00$\pm$0.00 \\
Computers    & 72  & 34 & 64 & 71 & 69 & 73  & 72.88$\pm$0.64  & 71.92$\pm$0.96  & 72.84$\pm$1.33  \\
CricketX     & 82  & 19 & 66 & 71 & 73 & 79  & 81.10$\pm$0.57  & 79.23$\pm$1.31  & 80.26$\pm$0.79  \\
CricketY     & 83  & 20 & 76 & 80 & 79 & 81  & 83.08$\pm$0.53  & 78.56$\pm$1.31  & 80.56$\pm$0.96  \\
CricketZ     & 83  & 62 & 73 & 80 & 81 & 82  & 82.62$\pm$0.32  & 82.54$\pm$0.57  & 82.74$\pm$0.44  \\
DiaSizRed    & 93  & 54 & 85 & 90 & 90 & 93  & 92.65$\pm$0.16  & 93.30$\pm$0.37  & 92.84$\pm$0.18  \\
DisPhaOAG    & 75  & 27 & 57 & 71 & 72 & 75  & 74.32$\pm$0.72  & 73.74$\pm$1.45  & 74.39$\pm$1.33  \\
DisPhaOutCor & 78  & 33 & 69 & 76 & 76 & 78  & 77.39$\pm$1.01  & 76.09$\pm$1.52  & 76.23$\pm$1.15  \\
DoLoDay      & 59  & 73 & 34 & 51 & 53 & 59  & 62.12$\pm$1.68  & 61.75$\pm$1.15  & 62.75$\pm$1.66  \\
DoLoGam      & 84  & 40 & 68 & 80 & 80 & 84  & 84.20$\pm$0.54  & 83.55$\pm$0.65  & 83.48$\pm$0.71  \\
DoLoWKE      & 98  & 1  & 90 & 95 & 95 & 97  & 98.55$\pm$0.00  & 98.55$\pm$0.00  & 98.55$\pm$0.00  \\
Earthquakes  & 75  & 1  & 59 & 73 & 74 & 75  & 74.82$\pm$0.00  & 74.82$\pm$0.00  & 75.25$\pm$1.03  \\
ECG200       & 92  & 20 & 84 & 89 & 90 & 91  & 91.70$\pm$0.46  & 91.00$\pm$1.10  & 91.10$\pm$0.70  \\
ECG5000      & 94  & 39 & 81 & 93 & 93 & 39  & 94.49$\pm$0.05  & 94.44$\pm$0.09  & 94.38$\pm$0.06  \\
ECGFiveDays  & 100 & 21 & 86 & 97 & 98 & 100 & 100.00$\pm$0.00 & 100.00$\pm$0.00 & 100.00$\pm$0.00 \\
EOGHSignal   & 60  & 29 & 46 & 56 & 57 & 59  & 59.78$\pm$1.12  & 56.19$\pm$1.38  & 58.15$\pm$1.18  \\
EOGVSignal   & 54  & 37 & 41 & 48 & 47 & 52  & 53.90$\pm$1.26  & 50.28$\pm$2.33  & 52.07$\pm$1.75  \\
FaceAll      & 81  & 72 & 71 & 79 & 79 & 81  & 80.65$\pm$0.17  & 80.88$\pm$0.17  & 80.79$\pm$0.12  \\
FaceFour     & 99  & 67 & 84 & 96 & 96 & 99  & 98.86$\pm$0.00  & 98.52$\pm$0.52  & 98.86$\pm$0.00  \\
FacesUCR     & 96  & 80 & 86 & 92 & 91 & 96  & 95.78$\pm$0.18  & 95.73$\pm$0.14  & 95.75$\pm$0.17  \\
FiftyWords   & 84  & 78 & 77 & 81 & 81 & 84  & 84.11$\pm$0.42  & 83.60$\pm$0.53  & 84.15$\pm$0.44 \\
\midrule
\textit{\textbf{AVERAGE} } &84 &36 &72 &80 &81   &83   &  84.37$\pm$0.46 & \textbf{83.48$\pm$0.83} & \textbf{84.12$\pm$0.66}\\
\bottomrule
\end{tabular}
\label{Minirocket_result}
\end{table*}

\subsection{Comparing with the Current State of the Art}
\label{Comparing with current state of the art}
\subsubsection{Settings}
In this section, we compare our {\cb POCKET} with the state-of-the-art pruning method for random kernels, S-ROCKET \cite{salehinejad2022s}. Following S-ROCKET, we prune the models for the first 30 datasets of the UCR archive.  
{\cb Referring to ROCKET and S-ROCKET,} here and in the rest of the experiments, we extend the range of values for $k$ in Stage 1 of {\cb POCKET} for validation to $[{\cb 0.01}, 0.1, 1, 10, 100, 1000]$ for a comprehensive investigation. All threads {\cb are used} to efficiently evaluate {\cb POCKET} and the compared methods.


As S-ROCKET is very time-consuming, {\cb when comparing with S-ROCKET on pruning ROCKET-PPV and MINIROCKET, we directly present the results reported in \cite{salehinejad2022s}.} For a fair comparison, we set the pruning rates and retraining settings in Stage 2 of our {\cb POCKET} the same as S-ROCKET.

To offer a more comprehensive evaluation, we also compare {\cb POCKET} and S-ROCKET {\cb on} pruning ROCKET-PPV-MAX. Since {\cb S-ROCKET had not pruned ROCKET-PPV-MAX}, we use its official code and reported settings for reproduction. For a fair comparison, {\cb POCKET}  adopts the pruning rates determined by S-ROCKET. However, when S-ROCKET produces an invalid pruning rate, i.e., when 100\% of the kernels remain, we eliminate 50\% of the random kernels. 



\begin{figure*}[htbp]
\centering
\subfigure[ROCKET-PPV-MAX]{
\includegraphics[width=0.315\linewidth]{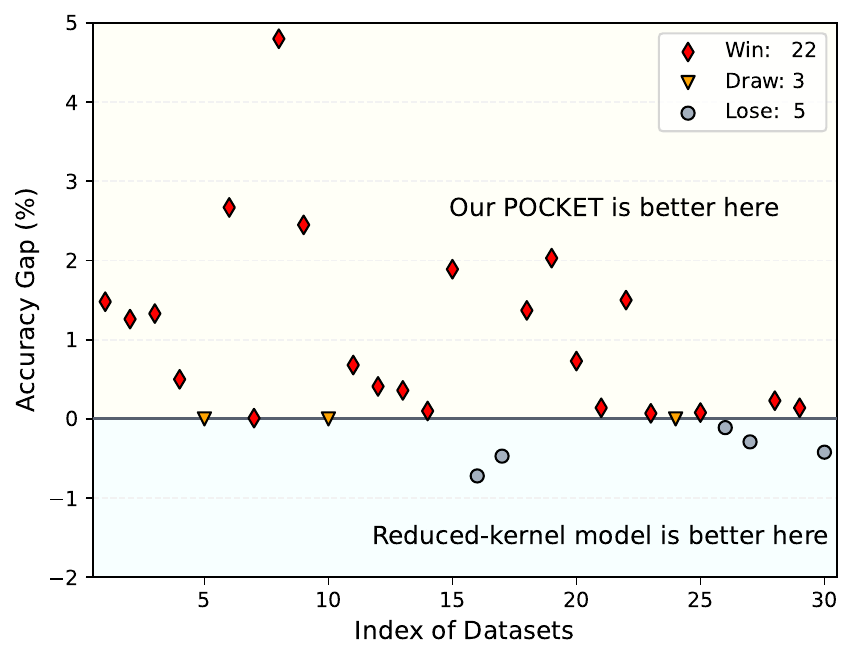}
}
\subfigure[ROCKET-PPV]{
\includegraphics[width=0.315\linewidth]{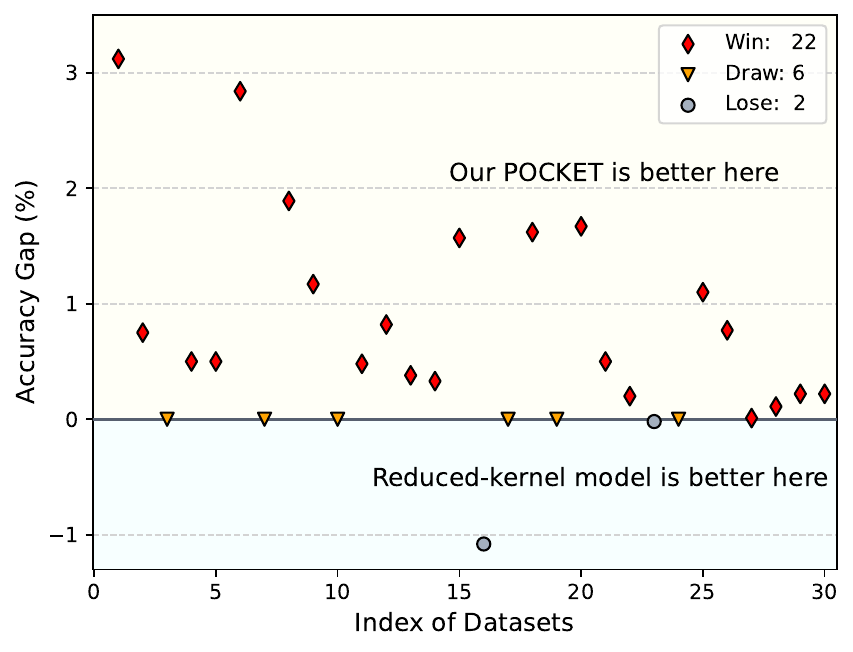}
}
\subfigure[MINIROCKET]{
\includegraphics[width=0.315\linewidth]{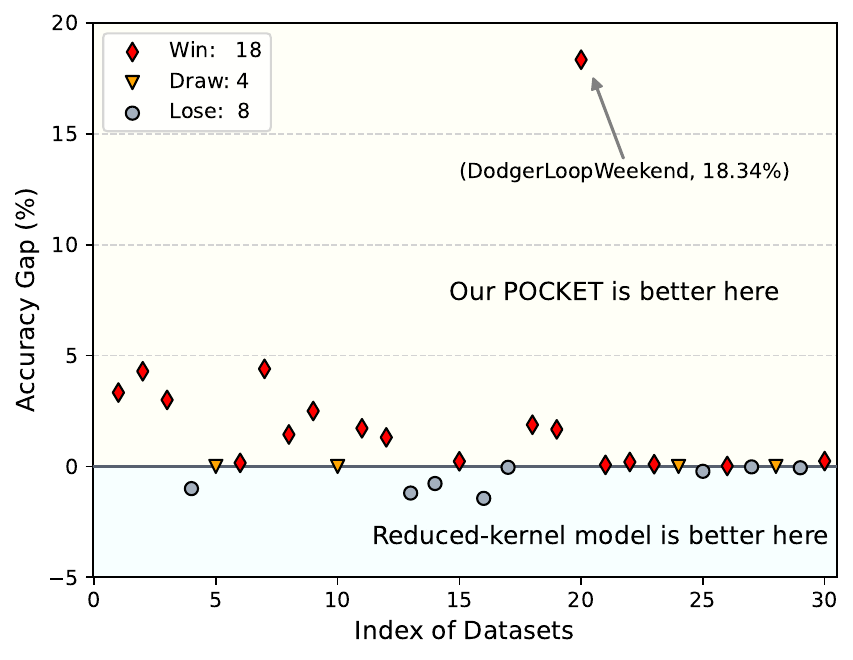}
}
\caption{Models pruned by {\cb POCKET} vs.  Reduced-kernel models trained from scratch.}
\label{VSFromScratch1}
\end{figure*}

\subsubsection{Results and Discussions}
\label{exp-b-2}
The results of pruning ROCKET-PPV-MAX and the training time comparison for different steps or stages between S-ROCKET and {\cb POCKET}  are presented in Table \ref{ROCKET_result} and \ref{Time}, respectively. Table \ref{ROCKET_PPV_result} and \ref{Minirocket_result} compare the pruned accuracy of ROCKET-PPV and MINIROCKET.
{\cb For comprehensive evaluation, we also report the overall results on pruning ROCKET-PPV-MAX across all 85 `bake off' and 43 `extra' UCR datasets in Table \ref{longtable1} and Table \ref{longtable12} in Appendix \ref{appendix1} and Appendix \ref{appendix2}, respectively.}

On average, {\cb POCKET} outperforms its counterparts by 0.5\% to 4\% in pruned accuracy across the three models, while being 11$\times$ faster than S-ROCKET.
Specifically, {\cbt despite removing more than 60\% of random kernels in ROCKET-PPV-MAX and ROCKET-PPV}, neither Stage 1 nor Stage 2  of {\cb POCKET} suffers from {\cbt severe} degradation; {\cbt they} even {\cbt achieve} higher average accuracy than the original unpruned models. 
The ability of POCKET {\cbt to reduce redundant features effectively decreases models' complexity}, {\cbt thereby alleviating} the overfitting issue {\cbt and} resulting in improved accuracy. {\cbt This improvement is also observed in deep learning compression \cite{chen2023whc,sun2020deep}, and we will further verify it} in the next subsection \ref{Effectiveness}. 

{\cbt As for MINIROCKET, since its capacity is limited by the use of discrete binary kernels,} pruning is more likely to cause significant degradation. Consequently, the pruned accuracy of MINIROCKET achieved in Stage 1 of POCKET decreases by 0.89\%. {\cbt However, Stage 2, which involves 
 $\ell_2$-regularized refitting,} significantly {\cbt enhances} POCKET's performance, {reducing} the gap between the original and pruned MINIROCKET to {\cbt just} 0.25\%.

In Table \ref{ROCKET_result}, while S-ROCKET \cite{salehinejad2022s} suffers from an invalid pruning rate in the `Adiac' dataset, our {\cb POCKET} eliminates redundancy with a pre-set rate, showing the notable advantage of {\cb POCKET} in enabling customized pruning rates. Additionally, although the MAX feature has been considered less important than the PPV \cite{dempster2021minirocket, salehinejad2022s}, it still contributes to a 1\% accuracy improvement for ROCKET-PPV-MAX after pruning, compared to ROCKET-PPV. This suggests that the importance of MAX may be underestimated.

{\cbt Comparing with S-ROCKET,}  the efficiency of Stage 1 enables {\cb POCKET} to operate 11$\times$ faster. However, Stage 1 costs a significant amount of time in cross-validation to determine the optimal value for $k$, indicating the necessity for more efficient parameter selection techniques.


\subsection{Effectiveness Verification}
\label{Effectiveness}

\subsubsection{Settings}
We verify the efficacy of {\cb POCKET} for pruning by comparing its pruned models with `\textbf{reduced-kernel models}', which refers to models trained from scratch with the number of kernels reduced to match the level achieved by {\cb POCKET} in the initialization phase, rather than the default 10K. The other settings for reduced-kernel models are the same as the original full models.
Without loss of generality, we directly adopt the pruning rates derived from S-ROCKET as outlined in Tables \ref{ROCKET_result}, \ref{ROCKET_PPV_result} and \ref{Minirocket_result} for the ROCKET-PPV-MAX, ROCKET-PPV and MINIROCKET models, respectively. 

\begin{figure}[htbp]
\centering
\includegraphics[width=1\linewidth]{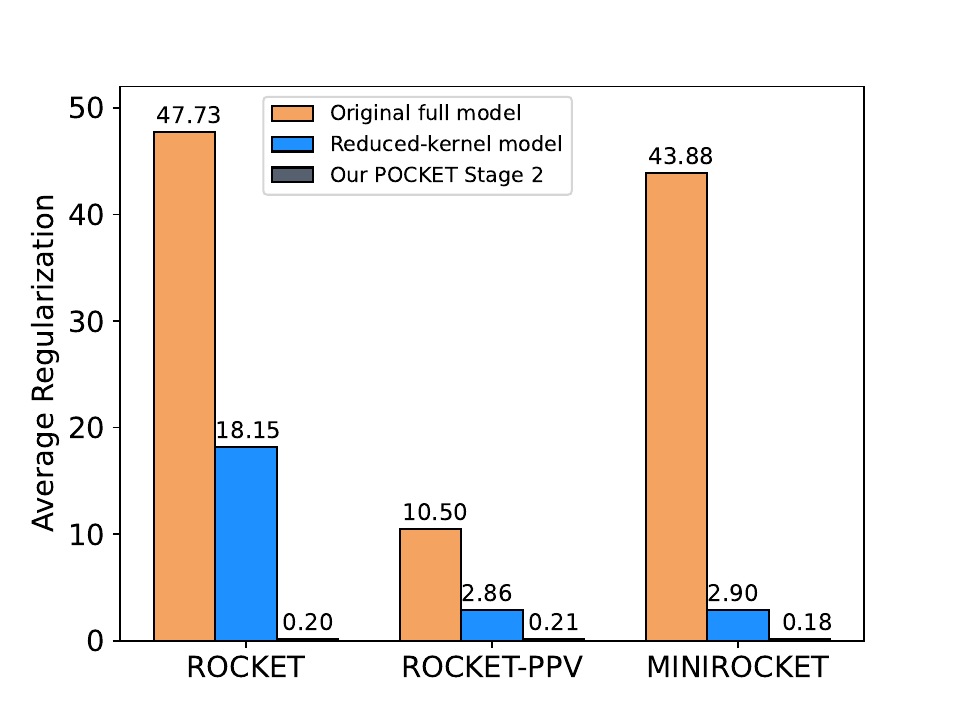}
\caption{Average $\ell_2$ regularization strength of models across datasets {\cb involved} in Figure \ref{VSFromScratch1}.}
\label{Regularization}
\vspace{-4mm}
\end{figure}

\begin{figure*}[htbp]
\centering
\subfigure[Adiac]{
\includegraphics[width=0.48\linewidth]{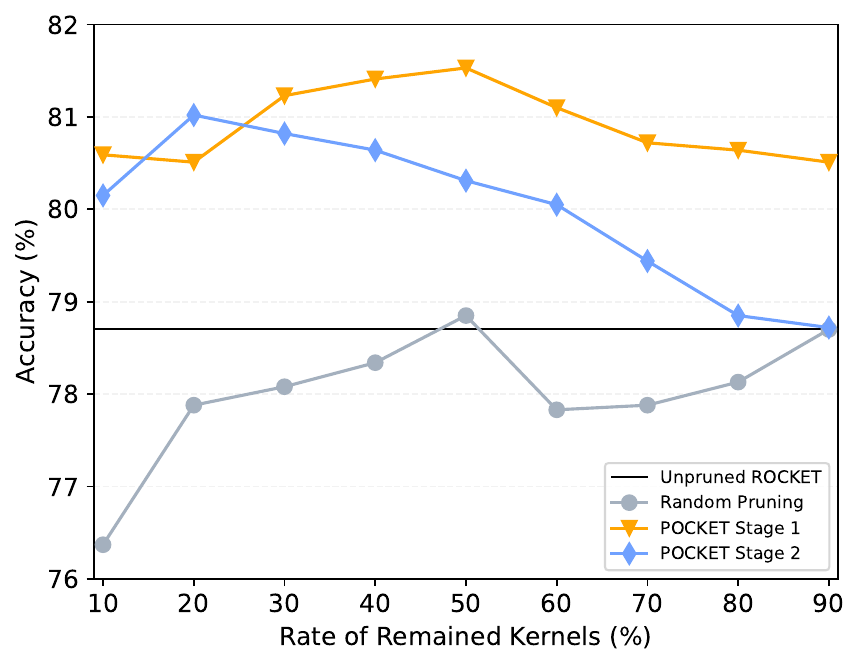}
}
\subfigure[ArrowHead]{
\includegraphics[width=0.48\linewidth]{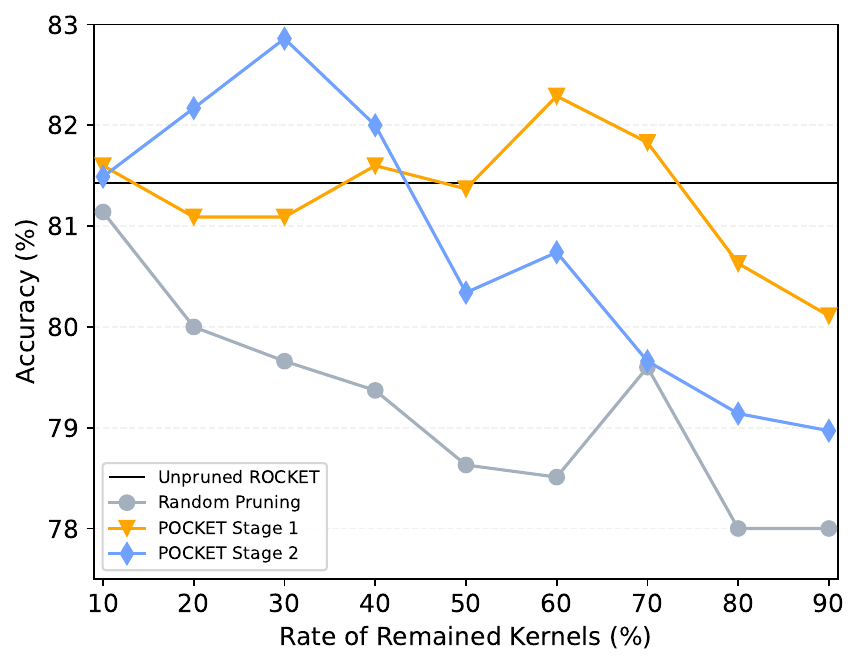}
}
\subfigure[Beef]{
\includegraphics[width=0.48\linewidth]{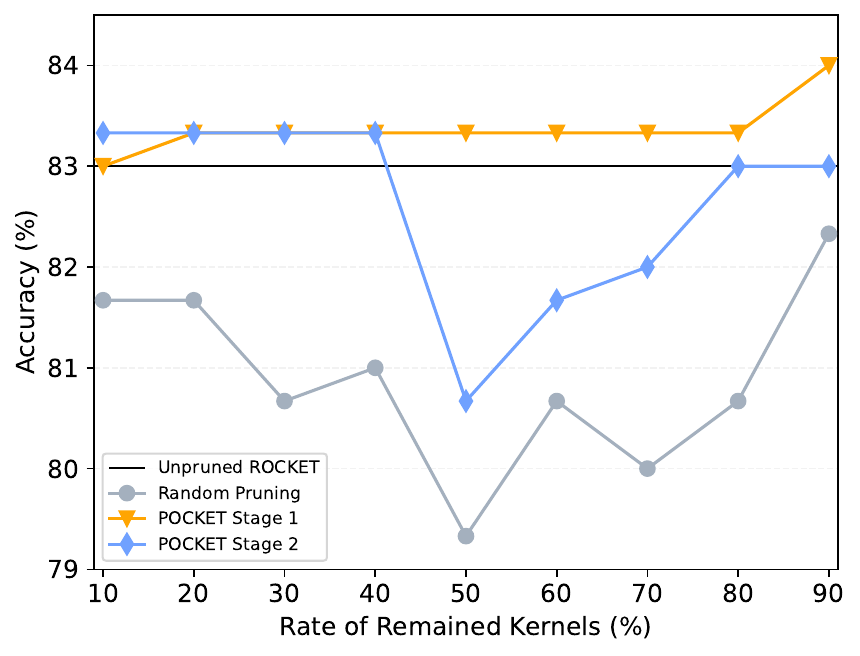}
}
\subfigure[BeetleFly]{
\includegraphics[width=0.48\linewidth]{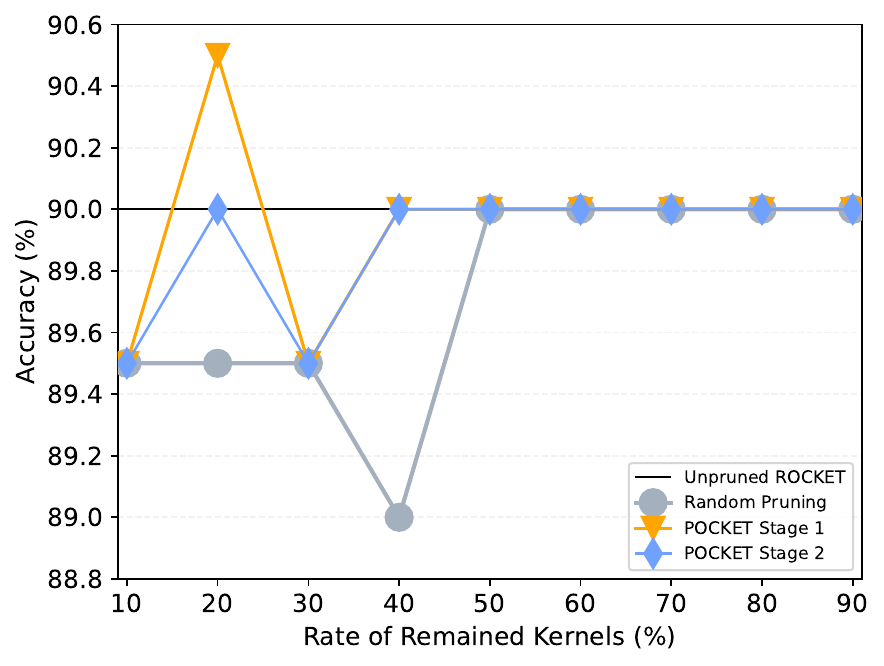}
}
\caption{Pruning ROCKET-PPV-MAX  for the first four datasets under different pruning rates.}
\label{PruningRate}
\vspace{-4mm}
\end{figure*}

\subsubsection{Results and Discussions}
As illustrated in Figure \ref{VSFromScratch1},  {\cb POCKET} consistently outperforms the reduced kernel models in most datasets, showing the efficacy of {\cb POCKET} in identifying crucial features to enhance performance.

In Figure \ref{Regularization}, we visualize the average regularization strength ($\rho_1$) of models across datasets. The strengths automatically selected by the efficient Leave-One-Out validation indicate the degree of redundancy. Not surprisingly, the original models (with full 10K kernels) have the strongest penalty strengths, followed by the reduced-kernel model. The models pruned by our {\cb POCKET} suffer the lightest penalty strength, suggesting that the remaining random kernels and features are crucial, while those pruned by {\cb POCKET} are indeed redundant. This observation also explains why models pruned by {\cb POCKET sometimes outperform the full models---the overfitting issue is mitigated by eliminating redundant parameters.}


\subsection{Pruning Rate}
\label{PruningRateSection}
\subsubsection{Settings}

An advantage of {\cb POCKET} is its adjustable pruning rate, which can be tailored to the computational budget. To {\cb evaluate POCKET} under different pruning rates, we prune 10\% to 90\% of random kernels on ROCKET-PPV-MAX and evaluate different stages of {\cb POCKET} across the first four datasets of the UCR archive. 
In comparison, we also assess Random Pruning under identical settings. Random Pruning {\cb removes kernels randomly}, providing a baseline for analyzing the impact of pruning rates.

\subsubsection{Results and Discussions}
The results are illustrated in Figure \ref{PruningRate}.
{\cb Generally, pruning kernels will cause degradation in accuracy, since the expression capacity of the model is reduced. 
However, as POCKET is efficient in recognizing redundancy, a relatively obvious improvement in accuracy, empirically a maximum of 3\%, can be achieved when 10\% to 30\% of kernels are removed.} Both Stage 1 and Stage 2 of {\cb POCKET} exhibit superior performance compared to random pruning {\cb with} a substantial margin across different pruning rates and datasets.
Notably, in the `Adiac' dataset, {\cb POCKET} outperforms the unpruned models at each pruning rate, underscoring its effectiveness in eliminating redundant features.

As discussed in \ref{exp-b-2}, the optional Stage 2 is less advantageous for ROCKET-PPV-MAX than for MINIROCKET in further enhancing accuracy. In the `Beef' dataset, Stage 2 exhibits similar trends to random pruning. Both Stage 2 and Random Pruning employ the same efficient Leave-one-Out validation approach, using the mean squared error as the evaluation metric instead of classification accuracy. From the perspective of classification accuracy, the selected values for hyperparameters may not be optimal, resulting in poor performance.  A more refined hyperparameter selection space and strategy could potentially ameliorate the situation, but would consume more computation time.

\subsection{More Datasets}
\label{MoreDatasets}
\subsubsection{Settings}
We proceed to comprehensively evaluate {\cb POCKET} by pruning ROCKET-PPV across all 128 datasets of the UCR archive. We keep only 10\% of random kernels for each dataset. Note that in the `Fungi' dataset, each class contains only one training sample, making cross-validation unavailable. To address this, we directly set the hyperparameter $k=1$. All other experimental settings remain consistent with those detailed in subsection \ref{General Settings}.

\subsubsection{Results and Discussions}

\begin{figure}[htbp]
\centering
\subfigure[{\cb POCKET} Stage 2 vs. reduced-kernel models from scratch]{
\includegraphics[width=1\linewidth]{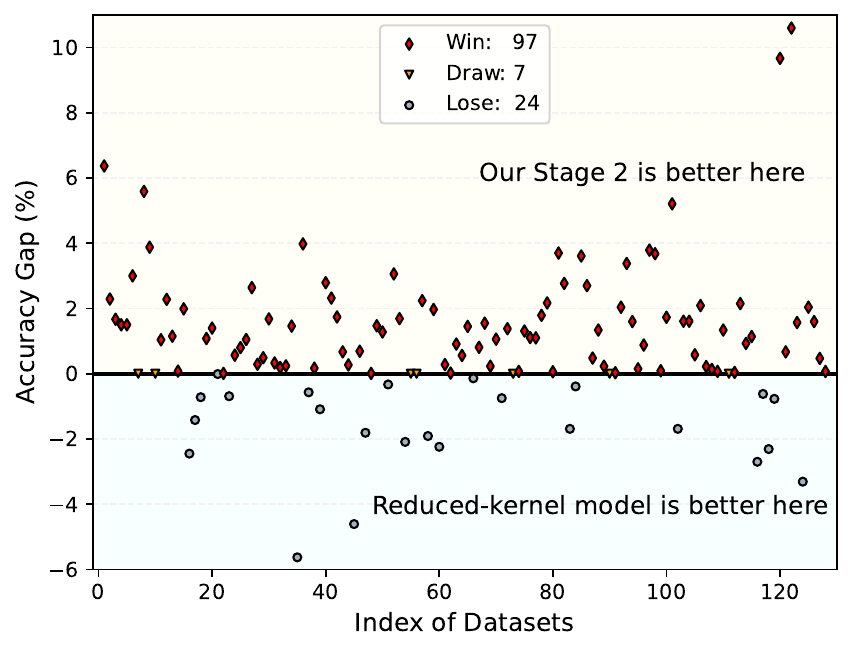}
}
\subfigure[{\cb POCKET} Stage 2 vs. {\cb POCKET} Stage 1]{
\includegraphics[width=1\linewidth]{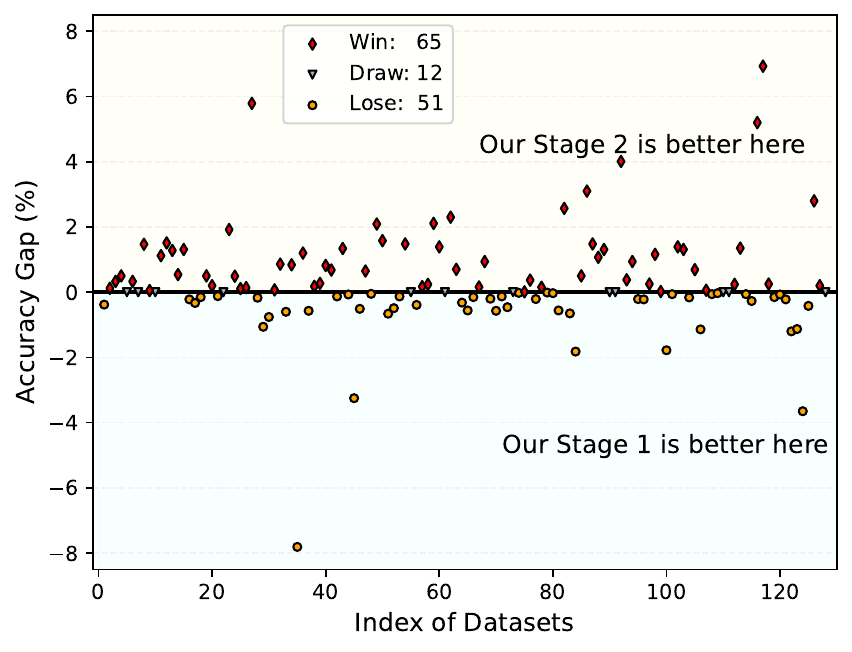}
}
\caption{Accuracy comparison across 128 UCR datasets when 90\% of kernels of ROCKET-PPV are pruned by POCKET.}
\label{allsets}
\vspace{-4mm}
\end{figure}

Figures \ref{allsets}-(a) and \ref{allsets}-(b) present comparisons between Stage 2 and reduced-kernel models, as well as Stage 1, respectively. Across the majority of datasets (97 out of 128), {\cb POCKET} demonstrates superior performance over the reduced-kernel ROCKET models. 

When comparing Stage 1 with Stage 2, the latter exhibits a slight performance improvement by a narrow margin on more datasets. In several specific instances (datasets indexed 35, 45, and 124), the reduced-kernel ROCKET models outperform Stage 2, while Stage 1 narrows the performance gaps. This aligns with the findings presented in the previous Subsection \ref{PruningRateSection}, where a non-optimal small penalty value selected through Leave-One-Out validation exacerbates the overfitting issue.  The nuanced performance dynamics between Stage 1 and Stage 2 underscore the importance of fine-tuning hyperparameters to strike the right balance between model complexity and overfitting across diverse datasets.

\begin{figure}[tbhp]
\centering
\includegraphics[width=1\linewidth]{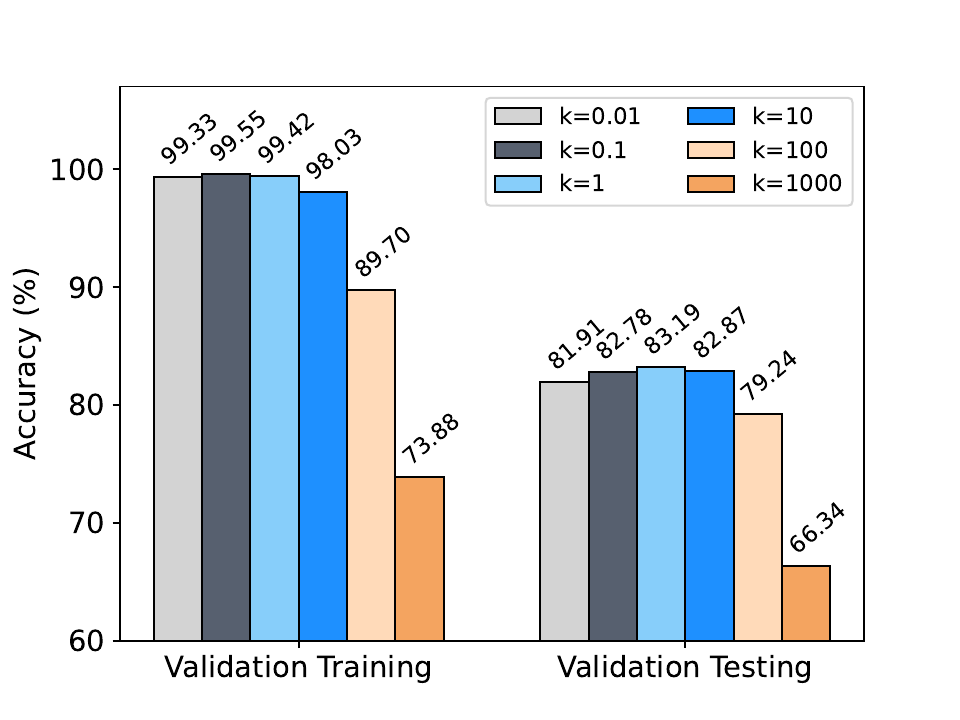}
\caption{{Average cross-validation accuracy of different} $k$ over 127 datasets.}
\label{k}
\vspace{-4mm}
\end{figure}\subsection{Hyperparameter $k$} 

\subsubsection{Settings}
To explore the impact of the hyperparameter $k$ on Stage 1 of {\cb POCKET}, we visualize the average cross-validation {\cb training and testing} accuracy {\cb over ten repetitions on 127 datasets,} excluding the `Fungi' dataset, {\cb whose $k$ is directly set to 1}. The experimental setup aligns with the parameters outlined in the previous experiment in \ref{MoreDatasets}.

\subsubsection{Results and Discussions}

The results presented in Figure \ref{k} indicate that the set \{0.1, 1, 10\} serves an appropriate range to achieve relatively higher cross-validation accuracy for both training and testing. Further refinement within the {\cb span} of [0.1, 10] {\cb has the potential} to enhance the performance of {\cb POCKET}. {\cb In addition,  the value $k=1$ is an appropriate choice for direct application without cross-validation.} {\cb Oppositely,}
the value $k=1000$ is {\cb too large, thereby} compromising classification accuracy across the majority of datasets.
Since a small $k$ prioritizes validation accuracy, while a large $k$ emphasizes group sparsity, the excessively large $k=1000$ {\cb results in performance degradation dramatically}.

\begin{figure*}[htbp]
\centering
\subfigure[Adiac]{
\includegraphics[width=0.315\linewidth]{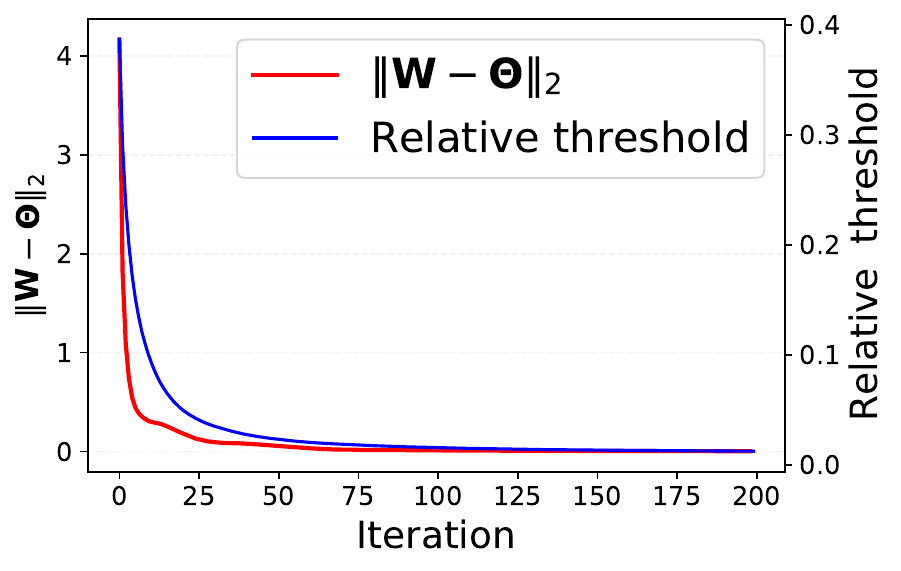}
}
\subfigure[ArrowHead]{
\includegraphics[width=0.315\linewidth]{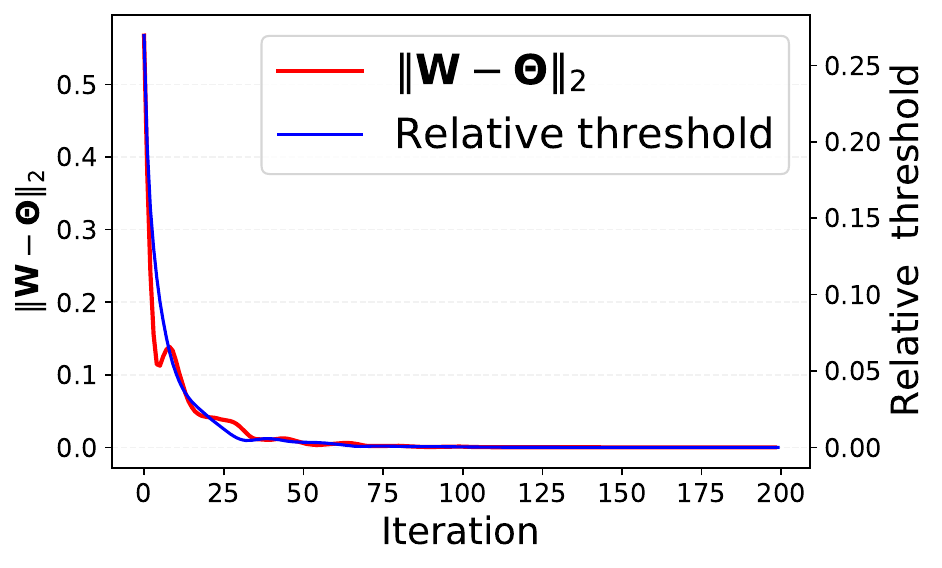}
}
\subfigure[Beef]{
\includegraphics[width=0.315\linewidth]{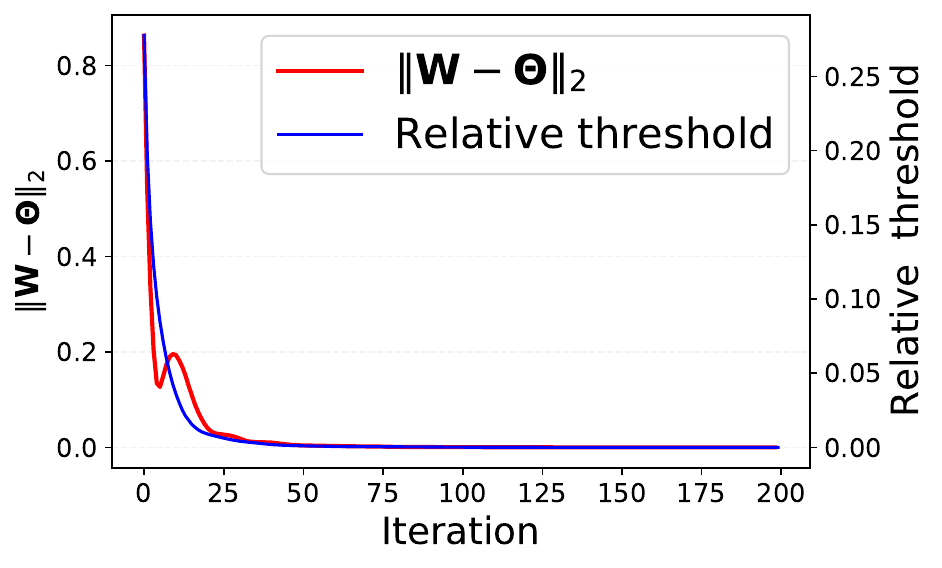}
}
\subfigure[BeetleFly]{
\includegraphics[width=0.315\linewidth]{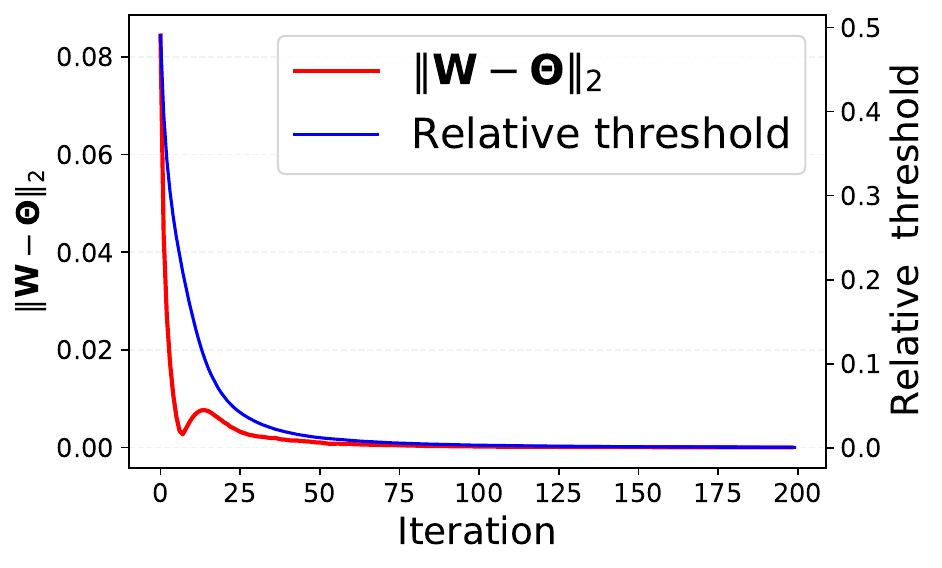}
}
\subfigure[BirdChicken]{
\includegraphics[width=0.315\linewidth]{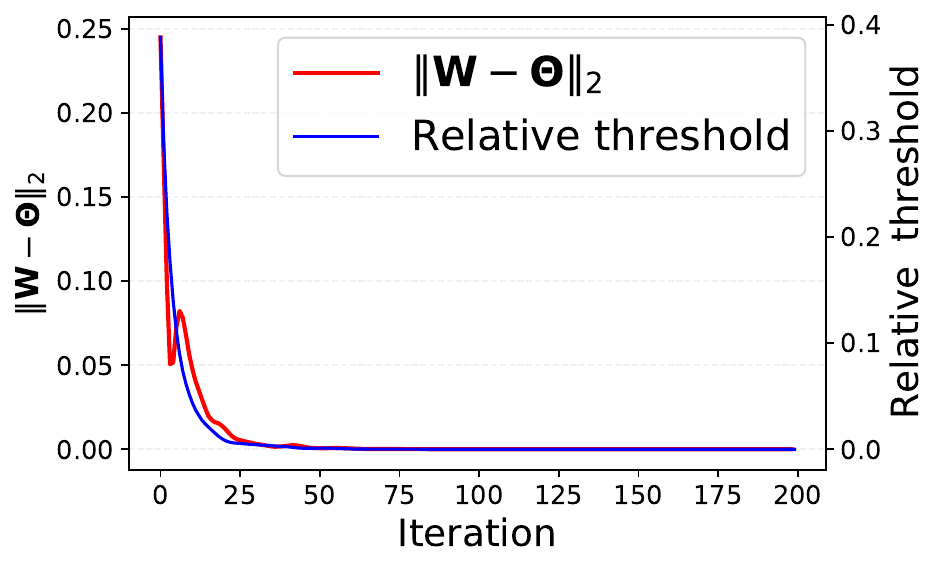}
}
\subfigure[Car]{
\includegraphics[width=0.315\linewidth]{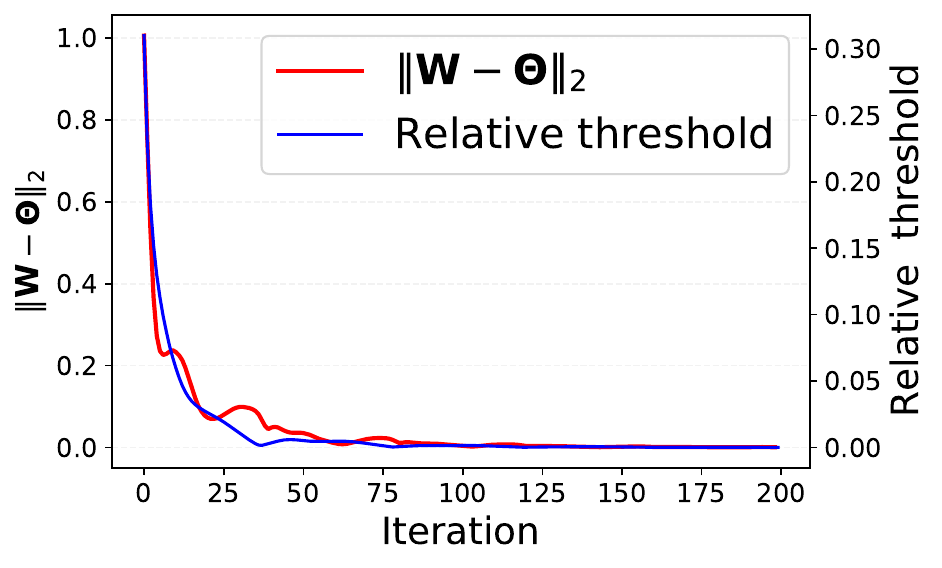}
}
\subfigure[CBF]{
\includegraphics[width=0.315\linewidth]{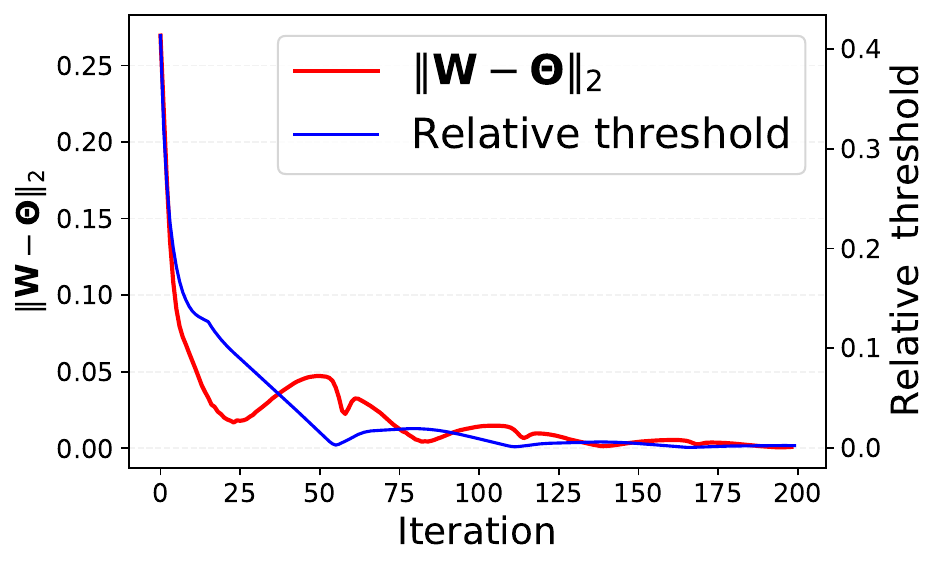}
}
\subfigure[ChlorineConcentration]{
\includegraphics[width=0.315\linewidth]{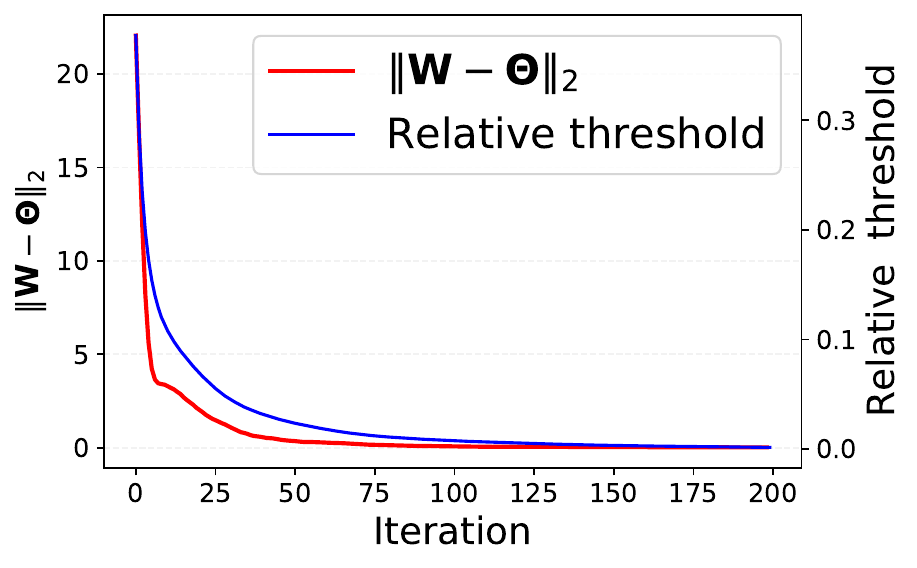}
}
\subfigure[CinCECGTorso]{
\includegraphics[width=0.315\linewidth]{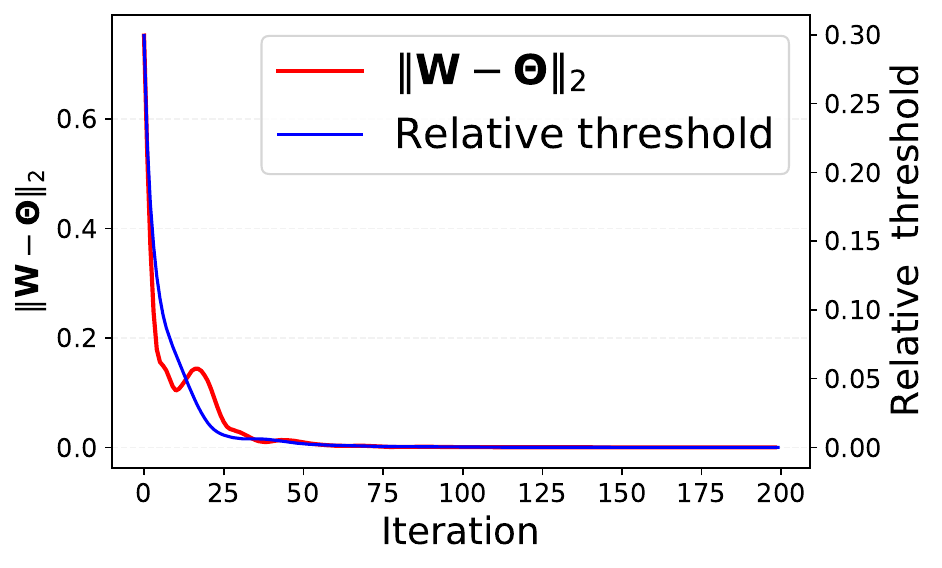}
}
\caption{Convergence process of Stage 1 in  {\cb POCKET} on pruning ROCKET-PPV. Here 90\% of the kernels are pruned. }
\label{convergence}
\vspace{-4mm}
\end{figure*}

\subsection{Convergence of Stage 1}
\subsubsection{Settings}
We visualize the convergence process of {\cb POCKET} Stage 1 across the first nine UCR datasets. The experimental settings are the same as in experiment \ref{MoreDatasets}, where 90\% of random kernels on ROCKET-PPV are pruned, except that the number of iterations is extended to 200 to comprehensively show the convergence process.
To avoid smoothing of convergence characteristics, we execute a single run for each dataset, rather than the ten repetitions.

\subsubsection{Results and Discussions}
In Figure \ref{convergence}, we plot the norm gap between $\W$ and $\bf{\Theta}$, i.e., $\Vert \W-\bf{\Theta}\Vert_2$. We also present the relative threshold, defined as $\frac{\Vert \widetilde{\W'}_{f_1:}^{(t)}\Vert_2}{\Vert \widetilde{\W'}_{f_{m+1}:}^{(t)}\Vert_2}$,  where $\Vert \widetilde{\W'}_{f_1:}^{(t)}\Vert_2$ and $\Vert \widetilde{\W'}_{f_{m+1}:}^{(t)}\Vert_2$ are defined in (\ref{rho_order}).
Using the relative threshold, as opposed to the direct threshold, mitigates the impact of the overall magnitude of the weights during iterations, providing a clearer illustration of the convergence.

It is shown that with an increasing number of iterations, the relative threshold progressively decreases towards 0, indicating a rapid convergence of $\W$ towards $\bTheta$.
Stage 1 of {\cb POCKET} achieves convergence in 50 to 200 iterations across most datasets, with notable instances {\cb such as `ArrowHead', `Beef', and `BirdChicken' where convergence is achieved in just 30 iterations.}
Considering the {\cb differing} convergence epochs observed {\cb in various} datasets, {\cb adaptive stopping criteria could potentially further expedite POCKET.} 
For example, {instead of fixing $T$, terminating} the iteration when {\cb the norm gap} falls below a predefined {\cb threshold could be more effective.}

\section{Conclusion}
\label{Conclusion}

{\cb This paper introduces two algorithms, the ADMM-based algorithm and the core POCKET, to efficiently prune models of the ROCKET family from a feature selection perspective,
thus avoiding the computational inefficiency caused by the direct evaluation of random kernels. Based on the ADMM-based algorithm, POCKET employs a two-stage strategy that significantly accelerates the pruning process, resulting in substantially faster speeds and higher accuracy compared with existing pruning methods.
Experimental results demonstrate that POCKET is effective in identifying redundant kernels and features. Specifically, on ROCKET-PPV-MAX, POCKET achieves an average kernel reduction exceeding 60\% without accuracy degradation, but improvements instead.}

In future work, we intend to {\cb integrate} the two classifiers generated {\cb during the} two stages of {\cb POCKET} for {\cb further enhancement in performance}, investigate more streamlined cross-validation techniques for higher efficiency, and {\cb explore the application of POCKET beyond time series classification.}

\printcredits

\section*{Acknowledgement}
This work was supported in part by the Guangdong Basic and Applied Basic Research Foundation under Grant 2021A1515011706, the National Science Fund for Distinguished Young Scholars under Grant 61925108, the Key Project of International Cooperation and Exchanges of the National Natural Science Foundation of China under Grant 62220106009, the project of Shenzhen Peacock Plan Teams under Grant KQTD20210811090051046, the Open Research Fund from the Guangdong Provincial Key Laboratory of Big Data Computing, the Chinese University of Hong Kong, Shenzhen, under Grant B10120210117-OF07, the Shenzhen University 2035 Program for Excellent Research under Grant 2022B009, the Research Team Cultivation Program of Shenzhen University under Grant 2023DFT003, and the Microelectronic Circuits Center Ireland under Grant MCCI-2020-07.

{\cb We extend our gratitude to Mr. Angus Dempster and Dr. Hojjat Salehinejad for their shared code and prompt email responses.}
We acknowledge the assistance of OpenAI’s language model, ChatGPT, in proofreading and enhancing the clarity of this manuscript. However, it is important to note that the generation of content lies solely with the authors.

\onecolumn

\appendix
{\cb

\section{85 `Bake off' UCR Datasets}
\label{appendix1}

\tablefirsthead{%
  \toprule
  \multirow{5}{*}{Dataset} &  {\cbt\multirow{5}{*}{\tabincell{l}{r-STSF\\ \cite{cabello2024fast}\\Acc.*\\(\%)}}
  } &{\cbt\multirow{5}{*}{\tabincell{l}{L-Time\\ \cite{LITE2023}\\Acc.*\\(\%)}}} & {\cbt\multirow{5}{*}{\tabincell{l}{H-ITime\\ \cite{H-IT2022}\\Acc.*\\(\%)}}}  & \multicolumn{8}{c}{\textbf{}ROCKET \& Pruning}\\
  \cmidrule(r){5-12}  
  &   &&&\multirow{4}{*}{{\cbt\tabincell{l}{{Unpruned}\\{ROCKET}\\{Acc.} \cite{dempster2020rocket}}}}&\multicolumn{3}{c}{{\cbt Pruned Models'} Acc. (\%)} &\multicolumn{2}{c}{$\#$Remaining Kernels} & \multicolumn{2}{c}{Overall Pruning Time (s)} \\
  \cmidrule(r){6-8}  \cmidrule(r){9-10} \cmidrule(r){11-12}
  &&&&    & \tabincell{l}{S-ROCKET\\ \cite{salehinejad2022s}} & \tabincell{l}{\textbf{POCKET}\\Stage 1} & \tabincell{l}{\textbf{POCKET}\\Stage 2}&\tabincell{l}{S-ROCKET\\ \cite{salehinejad2022s}} & \textbf{POCKET}&\tabincell{l}{S-ROCKET\\ \cite{salehinejad2022s}} & \textbf{POCKET}\\
  \midrule}
  
\tablehead{%
  \multicolumn{12}{l}{\small\sl continued from previous page} \\
  \toprule
  \multirow{5}{*}{Dataset} &  {\cbt\multirow{5}{*}{\tabincell{l}{r-STSF\\ \cite{cabello2024fast}\\Acc.*\\(\%)}}
  } &{\cbt\multirow{5}{*}{\tabincell{l}{L-Time\\ \cite{LITE2023}\\Acc.*\\(\%)}}} & {\cbt\multirow{5}{*}{\tabincell{l}{H-ITime\\ \cite{H-IT2022}\\Acc.*\\(\%)}}}  & \multicolumn{8}{c}{\textbf{}ROCKET \& Pruning}\\
  \cmidrule(r){5-12}  
  &   &&&\multirow{4}{*}{{\cbt\tabincell{l}{{Unpruned}\\{ROCKET}\\{Acc.} \cite{dempster2020rocket}}}}&\multicolumn{3}{c}{{\cbt Pruned Models'} Acc. (\%)} &\multicolumn{2}{c}{$\#$Remaining Kernels} & \multicolumn{2}{c}{Overall Pruning Time (s)} \\
  \cmidrule(r){6-8}  \cmidrule(r){9-10} \cmidrule(r){11-12}
  &&&&    & \tabincell{l}{S-ROCKET\\ \cite{salehinejad2022s}} & \tabincell{l}{\textbf{POCKET}\\Stage 1} & \tabincell{l}{\textbf{POCKET}\\Stage 2}&\tabincell{l}{S-ROCKET\\ \cite{salehinejad2022s}} & \textbf{POCKET}&\tabincell{l}{S-ROCKET\\ \cite{salehinejad2022s}} & \textbf{POCKET}\\
  \midrule}
  
\tabletail{%
  \midrule
  \multicolumn{12}{r}{\small\sl continued on next page} \\}
  
\tablelasttail{\bottomrule}

\begin{center}
\tablecaption{Performance comparison on pruning ROCKET-PPV-MAX across 85 `Bake off' UCR datasets. The columns marked with `*' are taken from the corresponding Github repository, while the other results are averaged over 10 repeated experiments. The datasets marked with `$^{+}$' are conducted on a Windows Server 2016 with two E5-2670 v3 CPUs, and the others are on a Ubuntu 18.04.6 Server with two Intel Xeon E5-2690 CPUs. `Acc.' stands for `Accuracy'; `\textbf{\textit{AVERAGE}}' represents the average over all listed datasets.
\label{longtable1}}
{\scriptsize
\setlength\tabcolsep{4pt}
\begin{supertabular}{llllllllllll}
\textit{\textbf{AVERAGE}}                    &  {\cbt 84.24}  & {\cbt 84.46}  & {\cbt 84.82}  & 85.04$\pm$0.51  & 84.88$\pm$0.73  & \color{red}{\textbf{85.17$\pm$0.84}} & 85.05$\pm$0.72  & 3861.64  & \color{red}{\textbf{3226.35}} & 3433.97  & \color{red}{\textbf{222.95}} \\
\midrule
Adiac            & {\cbt 83.55}  & {\cbt 83.38}  & {\cbt 84.40}  & 78.13$\pm$0.43  & 78.13$\pm$0.43  & 80.18$\pm$0.54  & 79.87$\pm$0.50  & 10000.00 & 5000.00 & 5551.44  & 270.69 \\
ArrowHead        & {\cbt 74.06}  & {\cbt 84.00}  & {\cbt 89.14}  & 81.37$\pm$1.03  & 81.77$\pm$1.31  & 80.86$\pm$1.93  & 81.83$\pm$1.59  & 2447.30  & 2447.30 & 398.68   & 136.33 \\
Beef             & {\cbt 87.00}  & {\cbt 76.67}  & {\cbt 70.00}  & 82.00$\pm$3.71  & 81.00$\pm$2.60  & 82.67$\pm$2.49  & 83.33$\pm$3.65  & 1945.80  & 1945.80 & 411.49   & 133.90 \\
BeetleFly        & {\cbt 92.00}  & {\cbt 90.00}  & {\cbt 75.00}  & 90.00$\pm$0.00  & 90.00$\pm$0.00  & 89.50$\pm$1.50  & 90.00$\pm$0.00  & 2107.80  & 2107.80 & 200.10   & 120.52 \\
BirdChicken      & {\cbt 90.50}  & {\cbt 90.00}  & {\cbt 95.00}  & 90.00$\pm$0.00  & 89.00$\pm$3.00  & 90.00$\pm$0.00  & 90.00$\pm$0.00  & 2430.50  & 2430.50 & 199.96   & 121.81 \\
Car              & {\cbt 87.50}  & {\cbt 93.33}  & {\cbt 90.00}  & 88.33$\pm$1.83  & 88.33$\pm$2.69  & 92.50$\pm$0.83  & 91.67$\pm$1.29  & 3428.50  & 3428.50 & 699.14   & 127.06 \\
CBF              & {\cbt 99.17}  & {\cbt 99.89}  & {\cbt 99.78}  & 100.00$\pm$0.00 & 99.96$\pm$0.05  & 99.92$\pm$0.07  & 99.98$\pm$0.04  & 1797.90  & 1797.90 & 344.33   & 137.40 \\
ChlCon           & {\cbt 77.82}  & {\cbt 84.84}  & {\cbt 88.26}  & 81.50$\pm$0.49  & 79.26$\pm$1.46  & 79.43$\pm$0.59  & 80.71$\pm$0.60  & 4066.90  & 4066.90 & 2228.87  & 162.91 \\
CinCECGTorso     & {\cbt 99.93}  & {\cbt 87.32}  & {\cbt 84.42}  & 83.61$\pm$0.55  & 82.79$\pm$0.74  & 90.86$\pm$2.96  & 88.23$\pm$1.64  & 2443.60  & 2443.60 & 482.42   & 133.08 \\
Coffee           & {\cbt 100.00} & {\cbt 100.00} & {\cbt 100.00} & 100.00$\pm$0.00 & 100.00$\pm$0.00 & 100.00$\pm$0.00 & 100.00$\pm$0.00 & 1805.60  & 1805.60 & 269.08   & 124.87 \\
Computers        & {\cbt 73.20}  & {\cbt 81.20}  & {\cbt 81.20}  & 76.32$\pm$0.84  & 76.80$\pm$0.95  & 74.60$\pm$1.82  & 77.20$\pm$0.76  & 2674.60  & 2674.60 & 1809.18  & 133.98 \\
CricketX         & {\cbt 74.51}  & {\cbt 83.08}  & {\cbt 85.38}  & 81.92$\pm$0.49  & 82.05$\pm$0.62  & 82.10$\pm$0.66  & 82.18$\pm$0.80  & 7294.70  & 7294.70 & 2821.87  & 167.17 \\
CricketY         & {\cbt 77.21}  & {\cbt 87.69}  & {\cbt 85.13}  & 85.38$\pm$0.60  & 85.08$\pm$0.56  & 83.87$\pm$0.75  & 84.90$\pm$0.71  & 5750.10  & 5250.10 & 2835.55  & 166.05 \\
CricketZ         & {\cbt 77.23}  & {\cbt 86.15}  & {\cbt 85.38}  & 85.44$\pm$0.64  & 85.03$\pm$0.69  & 83.87$\pm$0.71  & 85.10$\pm$0.71  & 7040.30  & 7040.30 & 2836.70  & 169.36 \\
DiaSizRed        & {\cbt 92.52}  & {\cbt 96.08}  & {\cbt 94.77}  & 97.09$\pm$0.61  & 96.50$\pm$0.84  & 95.59$\pm$2.37  & 97.84$\pm$0.55  & 2423.20  & 2423.20 & 209.88   & 124.90 \\
DisPhaOutAG      & {\cbt 73.02}  & {\cbt 71.22}  & {\cbt 74.10}  & 75.68$\pm$0.63  & 74.89$\pm$0.99  & 74.89$\pm$0.68  & 73.74$\pm$0.98  & 3284.80  & 3284.80 & 1899.12  & 146.55 \\
DisPhaOutCor     & {\cbt 78.12}  & {\cbt 76.09}  & {\cbt 79.71}  & 76.74$\pm$0.88  & 77.03$\pm$1.30  & 77.54$\pm$0.96  & 76.27$\pm$1.61  & 3271.00  & 3271.00 & 2499.27  & 146.57 \\
DisPhaTW$^+$     & {\cbt 68.06}  & {\cbt 71.22}  & {\cbt 65.47}  & 71.94$\pm$0.00  & 70.86$\pm$1.17  & 70.29$\pm$1.82  & 68.85$\pm$2.14  & 2389.40  & 2389.40 & 2013.88  & 123.28 \\
Earthquakes      & {\cbt 75.25}  & {\cbt 73.38}  & {\cbt 72.66}  & 74.82$\pm$0.00  & 74.82$\pm$0.00  & 74.96$\pm$0.29  & 74.96$\pm$0.54  & 3263.70  & 3263.70 & 2309.86  & 135.02 \\
ECG200           & {\cbt 89.70}  & {\cbt 92.00}  & {\cbt 90.00}  & 90.40$\pm$0.49  & 89.90$\pm$0.70  & 91.20$\pm$1.17  & 90.60$\pm$0.66  & 1088.10  & 1088.10 & 784.61   & 135.81 \\
ECG5000          & {\cbt 94.39}  & {\cbt 94.27}  & {\cbt 93.93}  & 94.75$\pm$0.05  & 94.68$\pm$0.09  & 93.55$\pm$0.60  & 94.78$\pm$0.06  & 3390.60  & 3390.60 & 2687.18  & 173.00 \\
ECGFiveDays      & {\cbt 99.48}  & {\cbt 99.77}  & {\cbt 100.00} & 100.00$\pm$0.00 & 100.00$\pm$0.00 & 100.00$\pm$0.00 & 100.00$\pm$0.00 & 2218.50  & 2218.50 & 205.96   & 124.82 \\
ElectricDev      & {\cbt 74.03}  & {\cbt 70.25}  & {\cbt 71.40}  & 72.81$\pm$0.25  & 72.72$\pm$0.39  & 72.65$\pm$0.38  & 72.63$\pm$0.30  & 4363.10  & 4363.10 & 48114.26 & 654.77 \\
FaceAll          & {\cbt 92.64}  & {\cbt 79.29}  & {\cbt 81.83}  & 94.68$\pm$0.40  & 94.64$\pm$0.32  & 94.16$\pm$0.67  & 94.14$\pm$0.78  & 4660.60  & 4660.60 & 4373.65  & 196.47 \\
FaceFour         & {\cbt 98.86}  & {\cbt 95.45}  & {\cbt 95.45}  & 97.61$\pm$0.34  & 97.61$\pm$0.34  & 98.41$\pm$0.56  & 97.84$\pm$0.34  & 1829.90  & 1829.90 & 321.47   & 125.82 \\
FacesUCR         & {\cbt 89.51}  & {\cbt 96.20}  & {\cbt 96.78}  & 96.20$\pm$0.09  & 96.20$\pm$0.08  & 96.14$\pm$0.16  & 96.34$\pm$0.08  & 4842.70  & 4842.70 & 1599.40  & 183.20 \\
FiftyWords       & {\cbt 76.99}  & {\cbt 80.66}  & {\cbt 84.84}  & 82.99$\pm$0.41  & 82.99$\pm$0.41  & 82.00$\pm$0.47  & 82.46$\pm$0.47  & 10000.00 & 5000.00 & 8068.53  & 308.62 \\
Fish             & {\cbt 92.91}  & {\cbt 97.71}  & {\cbt 98.29}  & 97.83$\pm$0.62  & 98.00$\pm$0.69  & 98.40$\pm$0.56  & 98.74$\pm$0.50  & 2257.90  & 1757.90 & 849.90   & 117.25 \\
FordA            & {\cbt 97.68}  & {\cbt 95.83}  & {\cbt 96.14}  & 94.43$\pm$0.28  & 94.04$\pm$0.26  & 94.61$\pm$0.18  & 94.05$\pm$0.57  & 2812.60  & 2812.60 & 13065.10 & 273.58 \\
FordB            & {\cbt 83.01}  & {\cbt 85.31}  & {\cbt 85.31}  & 80.43$\pm$0.78  & 79.81$\pm$0.39  & 80.54$\pm$0.36  & 80.52$\pm$0.67  & 3086.00  & 3086.00 & 14281.01 & 264.23 \\
GunPoint         & {\cbt 97.00}  & {\cbt 100.00} & {\cbt 100.00} & 100.00$\pm$0.00 & 100.00$\pm$0.00 & 99.33$\pm$0.00  & 100.00$\pm$0.00 & 1829.80  & 1829.80 & 314.20   & 104.12 \\
Ham              & {\cbt 76.95}  & {\cbt 71.43}  & {\cbt 70.48}  & 73.43$\pm$1.16  & 71.62$\pm$2.29  & 77.71$\pm$1.29  & 73.52$\pm$2.03  & 2292.30  & 2292.30 & 615.16   & 105.53 \\
HandOutlines     & {\cbt 91.57}  & {\cbt 95.41}  & {\cbt 95.41}  & 94.11$\pm$0.20  & 94.05$\pm$0.30  & 94.00$\pm$0.52  & 94.03$\pm$0.46  & 2704.70  & 2704.70 & 3849.30  & 152.11 \\
Haptics          & {\cbt 51.56}  & {\cbt 56.49}  & {\cbt 56.82}  & 52.11$\pm$0.36  & 52.31$\pm$0.75  & 52.50$\pm$0.70  & 52.89$\pm$0.91  & 3901.70  & 3901.70 & 742.73   & 131.02 \\
Herring          & {\cbt 60.47}  & {\cbt 76.56}  & {\cbt 71.88}  & 69.53$\pm$1.05  & 67.97$\pm$1.60  & 65.16$\pm$3.28  & 61.72$\pm$1.88  & 2863.80  & 2863.80 & 407.39   & 104.51 \\
InlineSkate      & {\cbt 66.75}  & {\cbt 55.09}  & {\cbt 52.55}  & 45.87$\pm$0.65  & 45.49$\pm$1.17  & 49.45$\pm$0.88  & 48.42$\pm$0.79  & 4943.00  & 2943.00 & 543.33   & 139.91 \\
InsWinSou        & {\cbt 66.80}  & {\cbt 63.33}  & {\cbt 64.44}  & 65.66$\pm$0.21  & 65.72$\pm$0.24  & 66.12$\pm$0.71  & 66.23$\pm$0.43  & 2763.50  & 2763.50 & 1354.89  & 133.89 \\
ItaPowDem        & {\cbt 97.31}  & {\cbt 96.70}  & {\cbt 96.89}  & 96.93$\pm$0.09  & 96.82$\pm$0.15  & 96.95$\pm$0.19  & 96.88$\pm$0.12  & 1050.70  & 1050.70 & 423.84   & 107.16 \\
LarKitApp        & {\cbt 80.64}  & {\cbt 89.07}  & {\cbt 89.60}  & 90.00$\pm$0.40  & 89.28$\pm$0.88  & 89.01$\pm$0.50  & 89.52$\pm$0.68  & 3535.60  & 3535.60 & 1613.64  & 135.65 \\
Lightning2       & {\cbt 76.72}  & {\cbt 75.41}  & {\cbt 80.33}  & 76.72$\pm$0.66  & 76.72$\pm$0.98  & 78.52$\pm$2.59  & 80.33$\pm$2.07  & 2626.50  & 2626.50 & 379.77   & 104.81 \\
Lightning7       & {\cbt 76.85}  & {\cbt 83.56}  & {\cbt 84.93}  & 82.19$\pm$0.61  & 82.88$\pm$1.10  & 80.96$\pm$1.88  & 82.47$\pm$1.19  & 3695.90  & 3695.90 & 797.78   & 111.92 \\
Mallat           & {\cbt 96.57}  & {\cbt 95.86}  & {\cbt 96.42}  & 95.63$\pm$0.21  & 95.63$\pm$0.21  & 92.99$\pm$1.09  & 95.52$\pm$0.26  & 10000.00 & 5000.00 & 694.15   & 120.00 \\
Meat             & {\cbt 95.00}  & {\cbt 93.33}  & {\cbt 95.00}  & 94.00$\pm$2.00  & 93.83$\pm$1.07  & 94.17$\pm$0.83  & 94.50$\pm$0.76  & 3599.30  & 3599.30 & 480.54   & 104.11 \\
MedicalImages    & {\cbt 81.67}  & {\cbt 78.55}  & {\cbt 80.13}  & 79.67$\pm$0.37  & 79.54$\pm$0.40  & 79.05$\pm$0.69  & 79.37$\pm$0.67  & 5141.30  & 4141.30 & 2278.48  & 131.79 \\
MidPhaOutAG      & {\cbt 59.35}  & {\cbt 51.30}  & {\cbt 51.95}  & 58.64$\pm$0.71  & 61.56$\pm$1.36  & 63.51$\pm$2.23  & 56.17$\pm$1.75  & 2795.10  & 2795.10 & 1707.16  & 114.49 \\
MidPhaOutCor     & {\cbt 83.61}  & {\cbt 84.88}  & {\cbt 83.85}  & 83.47$\pm$0.76  & 83.92$\pm$1.01  & 83.51$\pm$1.65  & 82.65$\pm$1.13  & 3098.90  & 3098.90 & 2288.76  & 118.32 \\
MiddlePTW        & {\cbt 59.68}  & {\cbt 50.00}  & {\cbt 51.30}  & 55.78$\pm$0.74  & 54.35$\pm$0.82  & 56.88$\pm$1.79  & 54.87$\pm$0.88  & 3797.90  & 3797.90 & 2012.32  & 119.74 \\
MoteStrain       & {\cbt 94.48}  & {\cbt 88.26}  & {\cbt 88.26}  & 91.41$\pm$0.30  & 91.53$\pm$0.44  & 91.71$\pm$0.29  & 91.38$\pm$0.38  & 1835.30  & 1835.30 & 144.22   & 105.41 \\
NonInvFECGT1$^+$ & {\cbt 93.64}  & {\cbt 95.98}  & {\cbt 96.08}  & 95.25$\pm$0.20  & 95.25$\pm$0.20  & 95.80$\pm$0.14  & 95.61$\pm$0.20  & 10000.00 & 5000.00 & 31179.69 & 596.24 \\
NonInvFECGT2$^+$ & {\cbt 94.60}  & {\cbt 96.08}  & {\cbt 96.69}  & 96.77$\pm$0.19  & 96.77$\pm$0.19  & 96.74$\pm$0.26  & 96.71$\pm$0.18  & 10000.00 & 5000.00 & 31973.74 & 599.28 \\
OliveOil$^+$     & {\cbt 90.00}  & {\cbt 80.00}  & {\cbt 76.67}  & 91.33$\pm$1.63  & 91.67$\pm$1.67  & 93.00$\pm$1.00  & 93.00$\pm$1.00  & 3194.50  & 3194.50 & 299.23   & 243.63 \\
OSULeaf$^+$      & {\cbt 84.88}  & {\cbt 95.45}  & {\cbt 95.04}  & 94.01$\pm$0.42  & 93.88$\pm$0.90  & 94.17$\pm$0.91  & 93.68$\pm$0.92  & 2544.80  & 2544.80 & 1003.20  & 253.27 \\
PhaOutCor$^+$    & {\cbt 84.06}  & {\cbt 84.03}  & {\cbt 84.27}  & 82.96$\pm$0.78  & 82.56$\pm$0.74  & 82.10$\pm$0.45  & 82.73$\pm$0.79  & 3258.50  & 3258.50 & 7441.26  & 308.97 \\
Phoneme$^+$      & {\cbt 39.76}  & {\cbt 33.70}  & {\cbt 33.81}  & 28.05$\pm$0.20  & 29.41$\pm$0.37  & 24.69$\pm$0.41  & 28.33$\pm$0.42  & 4732.60  & 4732.60 & 2607.88  & 389.57 \\
Plane$^+$        & {\cbt 100.00} & {\cbt 100.00} & {\cbt 100.00} & 100.00$\pm$0.00 & 100.00$\pm$0.00 & 100.00$\pm$0.00 & 100.00$\pm$0.00 & 7082.80  & 5582.80 & 577.74   & 251.66 \\
ProPhaOutAG$^+$  & {\cbt 85.41}  & {\cbt 85.37}  & {\cbt 83.90}  & 85.51$\pm$0.22  & 85.66$\pm$0.62  & 85.80$\pm$0.67  & 85.27$\pm$0.43  & 1705.40  & 1705.40 & 1752.40  & 243.38 \\
ProPhaOutCor$^+$ & {\cbt 92.06}  & {\cbt 91.75}  & {\cbt 92.44}  & 90.24$\pm$0.60  & 89.14$\pm$0.74  & 89.73$\pm$0.42  & 90.65$\pm$1.10  & 3019.40  & 3019.40 & 2312.24  & 248.79 \\
ProPhaTW$^+$     & {\cbt 79.12}  & {\cbt 80.49}  & {\cbt 78.05}  & 81.17$\pm$0.66  & 81.07$\pm$0.72  & 80.49$\pm$0.79  & 79.07$\pm$1.03  & 4931.30  & 3431.30 & 1998.84  & 254.29 \\
RefDev$^+$       & {\cbt 59.04}  & {\cbt 49.60}  & {\cbt 52.00}  & 53.49$\pm$0.96  & 53.20$\pm$0.87  & 54.08$\pm$0.73  & 53.41$\pm$1.23  & 3720.50  & 3720.50 & 1635.38  & 338.01 \\
ScreenType$^+$   & {\cbt 54.61}  & {\cbt 58.93}  & {\cbt 57.60}  & 48.56$\pm$1.15  & 49.20$\pm$2.09  & 46.99$\pm$1.54  & 47.92$\pm$0.97  & 3635.10  & 3635.10 & 1639.23  & 345.80 \\
ShapeletSim$^+$  & {\cbt 97.89}  & {\cbt 75.00}  & {\cbt 100.00} & 100.00$\pm$0.00 & 99.44$\pm$0.86  & 100.00$\pm$0.00 & 100.00$\pm$0.00 & 2084.20  & 2084.20 & 159.30   & 229.72 \\
ShapesAll$^+$    & {\cbt 86.12}  & {\cbt 91.33}  & {\cbt 92.83}  & 90.72$\pm$0.22  & 90.72$\pm$0.22  & 90.18$\pm$0.40  & 90.42$\pm$0.30  & 10000.00 & 5000.00 & 10350.97 & 446.24 \\
SmaKitApp$^+$    & {\cbt 82.35}  & {\cbt 77.07}  & {\cbt 77.60}  & 81.60$\pm$0.46  & 80.96$\pm$1.03  & 82.27$\pm$0.95  & 80.40$\pm$0.96  & 3184.10  & 3184.10 & 1638.70  & 348.32 \\
SonAIBORobS1$^+$ & {\cbt 89.58}  & {\cbt 83.53}  & {\cbt 84.86}  & 92.26$\pm$0.20  & 92.23$\pm$0.39  & 94.26$\pm$0.58  & 93.16$\pm$0.32  & 1821.50  & 1821.50 & 156.69   & 236.41 \\
SonAIBORobS2$^+$ & {\cbt 87.40}  & {\cbt 94.12}  & {\cbt 94.75}  & 91.22$\pm$0.27  & 91.22$\pm$0.52  & 92.13$\pm$0.36  & 92.24$\pm$0.55  & 3025.80  & 3025.80 & 181.42   & 241.38 \\
StarLightC$^+$   & {\cbt 97.94}  & {\cbt 97.67}  & {\cbt 97.56}  & 98.06$\pm$0.05  & 98.06$\pm$0.04  & 98.01$\pm$0.13  & 98.00$\pm$0.13  & 4035.80  & 2535.80 & 4544.64  & 409.38 \\
Strawberry$^+$   & {\cbt 96.84}  & {\cbt 98.38}  & {\cbt 98.38}  & 98.14$\pm$0.08  & 98.03$\pm$0.34  & 97.62$\pm$0.48  & 98.30$\pm$0.27  & 3239.80  & 3239.80 & 2399.56  & 260.15 \\
SwedishLeaf$^+$  & {\cbt 95.54}  & {\cbt 96.16}  & {\cbt 97.28}  & 96.56$\pm$0.29  & 96.32$\pm$0.35  & 96.45$\pm$0.34  & 96.50$\pm$0.27  & 8106.60  & 7106.60 & 3604.54  & 300.89 \\
Symbols$^+$      & {\cbt 97.24}  & {\cbt 98.29}  & {\cbt 97.99}  & 97.43$\pm$0.05  & 97.43$\pm$0.05  & 97.84$\pm$0.16  & 97.46$\pm$0.06  & 10000.00 & 5000.00 & 296.87   & 162.06 \\
SynCon$^+$       & {\cbt 99.00}  & {\cbt 100.00} & {\cbt 99.67}  & 99.97$\pm$0.10  & 99.73$\pm$0.29  & 99.03$\pm$0.10  & 99.83$\pm$0.22  & 5810.70  & 5810.70 & 1495.16  & 264.37 \\
ToeSeg1$^+$      & {\cbt 84.74}  & {\cbt 97.81}  & {\cbt 96.05}  & 96.80$\pm$0.34  & 96.62$\pm$0.34  & 95.26$\pm$1.25  & 95.88$\pm$0.35  & 1071.80  & 1071.80 & 267.47   & 244.17 \\
ToeSeg2$^+$      & {\cbt 87.92}  & {\cbt 93.85}  & {\cbt 93.08}  & 92.08$\pm$0.35  & 92.54$\pm$1.24  & 93.77$\pm$1.16  & 92.62$\pm$0.92  & 1091.60  & 1091.60 & 258.58   & 243.47 \\
Trace$^+$        & {\cbt 100.00} & {\cbt 100.00} & {\cbt 100.00} & 100.00$\pm$0.00 & 100.00$\pm$0.00 & 100.00$\pm$0.00 & 100.00$\pm$0.00 & 1826.30  & 1826.30 & 866.61   & 237.49 \\
TwoLeadECG$^+$   & {\cbt 98.44}  & {\cbt 99.82}  & {\cbt 99.74}  & 99.91$\pm$0.00  & 99.91$\pm$0.00  & 99.91$\pm$0.00  & 99.91$\pm$0.00  & 1840.70  & 1840.70 & 165.23   & 297.16 \\
TwoPatterns$^+$  & {\cbt 99.69}  & {\cbt 100.00} & {\cbt 100.00} & 100.00$\pm$0.00 & 100.00$\pm$0.00 & 100.00$\pm$0.00 & 100.00$\pm$0.00 & 1474.20  & 1474.20 & 4908.68  & 279.18 \\
UWavGesLibA$^+$  & {\cbt 95.59}  & {\cbt 91.23}  & {\cbt 94.78}  & 97.57$\pm$0.08  & 97.36$\pm$0.29  & 97.50$\pm$0.17  & 97.56$\pm$0.15  & 3914.80  & 3914.80 & 5160.15  & 362.54 \\
UWavGesLibX$^+$  & {\cbt 82.88}  & {\cbt 82.94}  & {\cbt 83.08}  & 85.50$\pm$0.21  & 85.28$\pm$0.47  & 85.14$\pm$0.42  & 85.26$\pm$0.28  & 6413.70  & 3413.70 & 5167.12  & 361.90 \\
UWavGesLibY$^+$  & {\cbt 75.75}  & {\cbt 76.16}  & {\cbt 78.28}  & 77.32$\pm$0.23  & 76.99$\pm$0.65  & 76.62$\pm$0.51  & 77.26$\pm$0.48  & 4142.30  & 2642.30 & 5208.30  & 362.11 \\
UWavGesLibZ$^+$  & {\cbt 76.81}  & {\cbt 78.08}  & {\cbt 78.53}  & 79.13$\pm$0.20  & 78.67$\pm$0.45  & 79.15$\pm$0.27  & 78.96$\pm$0.47  & 2373.20  & 1873.20 & 5176.96  & 361.32 \\
Wafer$^+$        & {\cbt 99.97}  & {\cbt 99.90}  & {\cbt 99.92}  & 99.83$\pm$0.01  & 99.78$\pm$0.06  & 99.86$\pm$0.04  & 99.83$\pm$0.06  & 1636.90  & 1636.90 & 4069.17  & 298.65 \\
Wine$^+$         & {\cbt 77.78}  & {\cbt 66.67}  & {\cbt 64.81}  & 80.93$\pm$2.99  & 80.37$\pm$3.72  & 83.15$\pm$5.46  & 82.96$\pm$4.96  & 2496.60  & 2496.60 & 364.49   & 240.98 \\
WordSynonyms$^+$ & {\cbt 65.39}  & {\cbt 71.79}  & {\cbt 75.86}  & 75.33$\pm$0.27  & 75.34$\pm$0.27  & 75.47$\pm$0.73  & 76.00$\pm$0.45  & 9825.40  & 5325.40 & 2422.28  & 317.08 \\
Worms$^+$        & {\cbt 79.22}  & {\cbt 81.82}  & {\cbt 83.12}  & 73.25$\pm$1.04  & 73.25$\pm$1.45  & 73.12$\pm$1.65  & 72.21$\pm$1.32  & 2563.70  & 2563.70 & 877.64   & 328.10 \\
WormsTwoCla$^+$  & {\cbt 80.52}  & {\cbt 76.62}  & {\cbt 77.92}  & 78.96$\pm$1.82  & 79.35$\pm$1.69  & 80.91$\pm$1.65  & 77.79$\pm$1.23  & 3403.80  & 3403.80 & 1026.81  & 324.07 \\
Yoga$^+$         & {\cbt 85.59}  & {\cbt 91.77}  & {\cbt 92.80}  & 91.16$\pm$0.36  & 90.43$\pm$0.43  & 91.38$\pm$0.43  & 91.48$\pm$0.52  & 2140.20  & 2140.20 & 1680.05  & 237.54 \\
\end{supertabular}}
\end{center}
\vspace{-20.8pt}

\section{ 43 `Extra'  UCR Datasets}
\label{appendix2}
\tablefirsthead{%
  \toprule
  \multirow{5}{*}{Dataset} &  {\cbt\multirow{5}{*}{\tabincell{l}{r-STSF\\ \cite{cabello2024fast}\\Acc.*\\(\%)}}
  } &{\cbt\multirow{5}{*}{\tabincell{l}{L-Time\\ \cite{LITE2023}\\Acc.*\\(\%)}}} & {\cbt\multirow{5}{*}{\tabincell{l}{H-ITime\\ \cite{H-IT2022}\\Acc.*\\(\%)}}}  & \multicolumn{8}{c}{\textbf{}ROCKET \& Pruning}\\
  \cmidrule(r){5-12}  
  &   &&&\multirow{4}{*}{{\cbt\tabincell{l}{{Unpruned}\\{ROCKET}\\{Acc.} \cite{dempster2020rocket}}}}&\multicolumn{3}{c}{{\cbt Pruned Models'} Acc. (\%)} &\multicolumn{2}{c}{$\#$Remaining Kernels} & \multicolumn{2}{c}{Overall Pruning Time (s)} \\
  \cmidrule(r){6-8}  \cmidrule(r){9-10} \cmidrule(r){11-12}
  &&&&    & \tabincell{l}{S-ROCKET\\ \cite{salehinejad2022s}} & \tabincell{l}{\textbf{POCKET}\\Stage 1} & \tabincell{l}{\textbf{POCKET}\\Stage 2}&\tabincell{l}{S-ROCKET\\ \cite{salehinejad2022s}} & \textbf{POCKET}&\tabincell{l}{S-ROCKET\\ \cite{salehinejad2022s}} & \textbf{POCKET}\\
  \midrule}

\tablehead{%
  \multicolumn{12}{l}{\small\sl continued from previous page} \\
  \toprule
  \multirow{5}{*}{Dataset} &  {\cbt\multirow{5}{*}{\tabincell{l}{r-STSF\\ \cite{cabello2024fast}\\Acc.*\\(\%)}}
  } &{\cbt\multirow{5}{*}{\tabincell{l}{L-Time\\ \cite{LITE2023}\\Acc.*\\(\%)}}} & {\cbt\multirow{5}{*}{\tabincell{l}{H-ITime\\ \cite{H-IT2022}\\Acc.*\\(\%)}}}  & \multicolumn{8}{c}{\textbf{}ROCKET \& Pruning}\\
  \cmidrule(r){5-12}  
  &   &&&\multirow{4}{*}{{\cbt\tabincell{l}{{Unpruned}\\{ROCKET}\\{Acc.} \cite{dempster2020rocket}}}}&\multicolumn{3}{c}{{\cbt Pruned Models'} Acc. (\%)} &\multicolumn{2}{c}{$\#$Remaining Kernels} & \multicolumn{2}{c}{Overall Pruning Time (s)} \\
  \cmidrule(r){6-8}  \cmidrule(r){9-10} \cmidrule(r){11-12}
  &&&&    & \tabincell{l}{S-ROCKET\\ \cite{salehinejad2022s}} & \tabincell{l}{\textbf{POCKET}\\Stage 1} & \tabincell{l}{\textbf{POCKET}\\Stage 2}&\tabincell{l}{S-ROCKET\\ \cite{salehinejad2022s}} & \textbf{POCKET}&\tabincell{l}{S-ROCKET\\ \cite{salehinejad2022s}} & \textbf{POCKET}\\
  \midrule}
  
\tabletail{%
  \midrule
  \multicolumn{12}{r}{\small\sl continued on next page} \\}
  
\tablelasttail{\bottomrule}

\begin{center}
\tablecaption{Performance comparison on pruning ROCKET-PPV-MAX across 43 `Extra' UCR datasets.  The columns marked with `*' are taken from the corresponding Github repository, while the other results are averaged over 10 repeated experiments. The datasets marked with `$^{+}$' are conducted on a Windows Server 2016 with two E5-2670 v3 CPUs, and the others are on a Ubuntu 18.04.6 Server with two Intel Xeon E5-2690 CPUs. `Acc.' stands for `Accuracy'; `\textbf{\textit{AVERAGE}}' represents the average over all listed datasets.\label{longtable12}}
{\scriptsize
\setlength\tabcolsep{3.8pt}

\begin{supertabular}{llllllllllll}
\textit{\textbf{AVERAGE}} & {\cbt-}  & {\cbt84.94} & {\cbt85.39}  & 85.18$\pm$0.48  & 84.90$\pm$0.72  & 85.29$\pm$1.03  & {\color{red} \textbf{85.48$\pm$0.78}} & 4492.17  &  {\color{red}\textbf{3573.57}} & 3312.54  &  {\color{red}\textbf{271.23}} \\
\midrule
ACSF1$^+$           & {\cbt 90.40}  & {\cbt 91.00}  & {\cbt 94.00}  & 87.90$\pm$0.83  & 87.30$\pm$1.35  & 89.80$\pm$1.33  & 89.40$\pm$1.74  & 6413.90  & 6413.90 & 614.10   & 257.82 \\
AllGesWiiX$^+$      & {\cbt -}      & {\cbt 76.57}  & {\cbt 79.14}  & 77.34$\pm$0.47  & 77.06$\pm$1.08  & 75.47$\pm$3.15  & 77.90$\pm$0.95  & 3218.20  & 3218.20 & 1740.51  & 257.72 \\
AllGesWiiY$^+$      & {\cbt -}      & {\cbt 78.86}  & {\cbt 83.57}  & 78.76$\pm$0.60  & 76.21$\pm$1.03  & 74.04$\pm$1.38  & 77.80$\pm$1.88  & 1353.20  & 1353.20 & 1740.39  & 259.80 \\
AllGesWiiZ$^+$      & {\cbt -}      & {\cbt 75.29}  & {\cbt 80.29}  & 76.69$\pm$0.50  & 76.00$\pm$0.93  & 70.71$\pm$3.41  & 75.49$\pm$1.06  & 2993.50  & 1993.50 & 1729.77  & 263.38 \\
BME$^+$             & {\cbt 99.87}  & {\cbt 99.33}  & {\cbt 99.33}  & 100.00$\pm$0.00 & 100.00$\pm$0.00 & 99.93$\pm$0.20  & 100.00$\pm$0.00 & 1854.50  & 1854.50 & 251.77   & 233.15 \\
Chinatown$^+$       & {\cbt 98.57}  & {\cbt 98.25}  & {\cbt 97.96}  & 98.25$\pm$0.00  & 98.25$\pm$0.18  & 98.25$\pm$0.00  & 98.51$\pm$0.09  & 2266.10  & 2266.10 & 157.63   & 225.23 \\
Crop$^+$            & {\cbt 77.70}  & {\cbt 75.73}  & {\cbt 77.21}  & 76.64$\pm$0.07  & 75.85$\pm$1.27  & 73.21$\pm$1.92  & 75.26$\pm$0.74  & 1883.40  & 1383.40 & 66160.18 & 826.16 \\
DodgerLoopDay       & {\cbt -}      & {\cbt 57.50}  & {\cbt 58.75}  & 60.50$\pm$1.70  & 60.25$\pm$1.84  & 60.88$\pm$1.68  & 60.62$\pm$2.11  & 5520.20  & 5020.20 & 512.74   & 154.22 \\
DodgerLoopGame      & {\cbt -}      & {\cbt 81.88}  & {\cbt 83.33}  & 86.09$\pm$0.54  & 86.45$\pm$0.80  & 88.33$\pm$0.68  & 88.26$\pm$0.63  & 1842.60  & 1842.60 & 199.85   & 123.50 \\
DodLooWee           & {\cbt -}      & {\cbt 97.10}  & {\cbt 97.10}  & 97.68$\pm$0.29  & 97.54$\pm$0.35  & 98.55$\pm$0.00  & 98.48$\pm$0.22  & 1827.50  & 1827.50 & 199.72   & 121.82 \\
EOGHorSig           & {\cbt 57.15}  & {\cbt 63.54}  & {\cbt 62.15}  & 58.26$\pm$1.05  & 58.01$\pm$1.19  & 59.17$\pm$1.46  & 58.45$\pm$1.18  & 6933.10  & 6433.10 & 2627.53  & 168.83 \\
EOGVerSig           & {\cbt 52.82}  & {\cbt 41.16}  & {\cbt 46.69}  & 54.70$\pm$0.58  & 55.03$\pm$0.91  & 55.28$\pm$0.94  & 54.81$\pm$0.58  & 5156.00  & 5156.00 & 2635.61  & 165.08 \\
EthanolLevel$^+$    & {\cbt 61.84}  & {\cbt 74.00}  & {\cbt 78.80}  & 58.56$\pm$0.84  & 57.40$\pm$0.80  & 58.86$\pm$1.38  & 59.76$\pm$0.92  & 2899.80  & 2899.80 & 2651.43  & 363.85 \\
FreRegTra$^+$       & {\cbt 99.92}  & {\cbt 99.79}  & {\cbt 99.72}  & 99.76$\pm$0.03  & 99.69$\pm$0.07  & 99.75$\pm$0.04  & 99.77$\pm$0.03  & 2959.60  & 2959.60 & 1094.73  & 250.27 \\
FreSmaTra$^+$       & {\cbt 98.14}  & {\cbt 96.11}  & {\cbt 82.25}  & 95.14$\pm$0.18  & 94.92$\pm$0.64  & 95.20$\pm$3.17  & 96.24$\pm$0.40  & 1833.70  & 1833.70 & 239.96   & 238.99 \\
Fungi$^+$           & {\cbt -}      & {\cbt 100.00} & {\cbt 100.00} & 99.41$\pm$0.16  & 99.41$\pm$0.16  & 100.00$\pm$0.00 & 99.78$\pm$0.26  & 9756.90  & 5256.90 & 409.52   & 37.10  \\
GestureMidAirD1$^+$ & {\cbt -}      & {\cbt 75.38}  & {\cbt 75.38}  & 77.85$\pm$0.58  & 77.85$\pm$0.58  & 74.92$\pm$1.38  & 77.62$\pm$0.80  & 10000.00 & 5000.00 & 1898.31  & 345.22 \\
GestureMidAirD2$^+$ & {\cbt -}      & {\cbt 70.77}  & {\cbt 73.08}  & 66.77$\pm$1.02  & 66.77$\pm$1.02  & 68.62$\pm$1.37  & 67.15$\pm$0.60  & 10000.00 & 5000.00 & 1898.97  & 335.66 \\
GestureMidAirD3$^+$ & {\cbt -}      & {\cbt 42.31}  & {\cbt 42.31}  & 39.62$\pm$0.62  & 39.69$\pm$0.62  & 39.69$\pm$2.13  & 40.54$\pm$0.98  & 9768.60  & 5268.60 & 1900.54  & 311.46 \\
GesturePebbleZ1$^+$ & {\cbt -}      & {\cbt 91.86}  & {\cbt 93.02}  & 90.17$\pm$0.31  & 90.29$\pm$0.52  & 90.00$\pm$0.35  & 90.23$\pm$0.35  & 4825.90  & 4825.90 & 683.48   & 255.97 \\
GesturePebbleZ2$^+$ & {\cbt -}      & {\cbt 89.87}  & {\cbt 88.61}  & 83.73$\pm$0.85  & 83.61$\pm$1.11  & 84.68$\pm$1.05  & 84.56$\pm$0.95  & 7076.70  & 7076.70 & 1185.65  & 263.11 \\
GunPointAgeSpan$^+$ & {\cbt 98.99}  & {\cbt 99.05}  & {\cbt 99.68}  & 99.02$\pm$0.22  & 98.61$\pm$0.60  & 97.12$\pm$1.46  & 98.80$\pm$0.37  & 1791.10  & 1791.10 & 1476.98  & 243.42 \\
GunPoiMalVerFem$^+$ & {\cbt 100.00} & {\cbt 99.05}  & {\cbt 100.00} & 99.78$\pm$0.20  & 99.75$\pm$0.19  & 100.00$\pm$0.00 & 99.91$\pm$0.15  & 963.80   & 963.80  & 1475.11  & 239.70 \\
GunPoiOldVerYou$^+$ & {\cbt 100.00} & {\cbt 97.78}  & {\cbt 97.78}  & 100.00$\pm$0.00 & 100.00$\pm$0.00 & 100.00$\pm$0.00 & 100.00$\pm$0.00 & 1825.90  & 1825.90 & 1465.64  & 246.85 \\
HouseTwenty$^+$     & {\cbt 91.93}  & {\cbt 98.32}  & {\cbt 97.48}  & 96.30$\pm$0.41  & 96.30$\pm$0.41  & 96.30$\pm$0.41  & 95.97$\pm$0.50  & 2517.30  & 2517.30 & 523.17   & 244.51 \\
InsEPGRegTra$^+$    & {\cbt 100.00} & {\cbt 99.60}  & {\cbt 100.00} & 100.00$\pm$0.00 & 99.96$\pm$0.12  & 100.00$\pm$0.00 & 100.00$\pm$0.00 & 2114.60  & 2114.60 & 972.76   & 246.39 \\
InsEPGSmaTra$^+$    & {\cbt 100.00} & {\cbt 96.39}  & {\cbt 91.57}  & 100.00$\pm$0.00 & 99.88$\pm$0.36  & 100.00$\pm$0.00 & 100.00$\pm$0.00 & 2114.40  & 2114.40 & 326.46   & 246.87 \\
MelPed$^+$          & {\cbt -}      & {\cbt 90.57}  & {\cbt 91.18}  & 94.46$\pm$0.23  & 94.16$\pm$0.35  & 94.09$\pm$0.41  & 94.12$\pm$0.33  & 4981.40  & 4981.40 & 9637.49  & 311.78 \\
MixShaRegTra$^+$    & {\cbt 94.90}  & {\cbt 98.06}  & {\cbt 97.65}  & 97.06$\pm$0.09  & 97.01$\pm$0.11  & 97.13$\pm$0.13  & 97.02$\pm$0.11  & 3449.90  & 3449.90 & 3474.85  & 405.09 \\
MixShaSmaTra$^+$    & {\cbt 89.59}  & {\cbt 94.39}  & {\cbt 92.16}  & 93.78$\pm$0.09  & 93.55$\pm$0.26  & 93.88$\pm$0.19  & 93.58$\pm$0.12  & 2829.80  & 2829.80 & 1821.62  & 278.92 \\
PicGesWiiZ$^+$      & {\cbt -}      & {\cbt 78.00}  & {\cbt 78.00}  & 79.40$\pm$0.92  & 79.40$\pm$0.92  & 82.20$\pm$1.08  & 81.80$\pm$1.66  & 10000.00 & 5000.00 & 1404.72  & 284.39 \\
PigAirPre$^+$       & {\cbt 37.50}  & {\cbt 47.12}  & {\cbt 56.73}  & 80.05$\pm$0.72  & 80.05$\pm$0.72  & 84.71$\pm$1.23  & 84.86$\pm$0.62  & 10000.00 & 5000.00 & 2412.84  & 248.78 \\
PigArtPressure$^+$  & {\cbt 92.40}  & {\cbt 99.52}  & {\cbt 99.52}  & 95.77$\pm$0.47  & 95.77$\pm$0.47  & 93.17$\pm$0.52  & 94.81$\pm$0.77  & 8835.50  & 5335.50 & 2374.94  & 232.38 \\
PigCVP$^+$          & {\cbt 68.32}  & {\cbt 94.71}  & {\cbt 96.63}  & 93.12$\pm$0.22  & 93.12$\pm$0.22  & 92.98$\pm$0.24  & 93.03$\pm$0.32  & 9764.00  & 5264.00 & 2383.67  & 230.76 \\
PLAID$^+$           & {\cbt -}      & {\cbt 91.99}  & {\cbt 94.23}  & 81.38$\pm$1.51  & 80.24$\pm$2.05  & 84.00$\pm$2.25  & 83.13$\pm$1.67  & 3675.00  & 3675.00 & 4881.03  & 373.56 \\
PowerCons$^+$       & {\cbt 100.00} & {\cbt 96.11}  & {\cbt 93.89}  & 98.56$\pm$0.44  & 98.56$\pm$1.03  & 99.83$\pm$0.50  & 99.28$\pm$0.70  & 784.10   & 784.10  & 1921.19  & 250.39 \\
Rock$^+$            & {\cbt 75.80}  & {\cbt 86.00}  & {\cbt 84.00}  & 68.20$\pm$0.60  & 69.00$\pm$1.84  & 80.00$\pm$1.79  & 73.00$\pm$3.00  & 4032.00  & 4032.00 & 411.11   & 256.21 \\
SemHanGenCh2$^+$    & {\cbt 96.60}  & {\cbt 86.83}  & {\cbt 82.33}  & 89.45$\pm$0.62  & 88.40$\pm$0.59  & 88.52$\pm$0.80  & 88.47$\pm$0.60  & 3053.50  & 3053.50 & 3310.66  & 352.21 \\
SemHanMovCh2$^+$    & {\cbt 83.49}  & {\cbt 51.11}  & {\cbt 53.11}  & 59.64$\pm$1.06  & 57.76$\pm$1.42  & 56.64$\pm$3.23  & 55.09$\pm$3.02  & 2924.80  & 2924.80 & 3785.82  & 369.80 \\
SemHanSubCh2$^+$    & {\cbt 88.93}  & {\cbt 80.22}  & {\cbt 84.44}  & 84.67$\pm$0.82  & 83.33$\pm$0.92  & 82.51$\pm$1.54  & 83.27$\pm$1.39  & 4115.40  & 4115.40 & 3307.84  & 358.55 \\
ShaGesWiiZ$^+$      & {\cbt -}      & {\cbt 96.00}  & {\cbt 92.00}  & 92.20$\pm$0.60  & 92.00$\pm$0.89  & 93.80$\pm$1.08  & 92.40$\pm$1.20  & 7122.70  & 7122.70 & 1475.76  & 278.25 \\
SmoothSubspace$^+$  & {\cbt 98.00}  & {\cbt 98.67}  & {\cbt 97.33}  & 97.47$\pm$0.27  & 97.27$\pm$0.55  & 96.73$\pm$0.36  & 97.67$\pm$0.33  & 3084.50  & 3084.50 & 2391.30  & 254.11 \\
UMD$^+$             & {\cbt 99.79}  & {\cbt 96.53}  & {\cbt 99.31}  & 98.61$\pm$0.00  & 98.82$\pm$0.32  & 98.61$\pm$0.00  & 98.61$\pm$0.00  & 2800.30  & 2800.30 & 671.92   & 251.52\\

\end{supertabular}}
\end{center}
}

\twocolumn

\bibliographystyle{model1-num-names}

\bibliography{ref}






\end{document}